%% file: neurips_2026.tex
\documentclass{article}

\PassOptionsToPackage{numbers, compress}{natbib}

\usepackage[eandd, final]{neurips_2026}

\usepackage[utf8]{inputenc} 
\usepackage[T1]{fontenc}    
\usepackage{url}            
\usepackage{booktabs}       
\usepackage{tabularx}
\usepackage{amsfonts}       
\usepackage{nicefrac}       
\usepackage{microtype}      
\usepackage{xcolor}         
\usepackage{graphicx}
\usepackage{enumitem}
\usepackage{amsmath}
\usepackage{wrapfig}        
\usepackage{gradient-text}
\usepackage[most]{tcolorbox}
\usepackage{setspace}
\usepackage{multirow}
\usepackage{algorithm}    
\usepackage{algpseudocode}
\usepackage{colortbl}
\usepackage{soul}
\usepackage{caption}
\usepackage{makecell}
\usepackage[normalem]{ulem}

\definecolor{blue}{rgb}{0.21,0.49,0.74}
\usepackage[colorlinks=true, linkcolor=blue, citecolor=blue, urlcolor=blue]{hyperref}
\usepackage[capitalise]{cleveref} 

\title{nuReasoning: A Reasoning-Centric Dataset and Benchmark for Long-Tail Autonomous Driving}

\author{
  Zhiyu Huang$^{1}$\thanks{Equal contribution.} \quad
  Johnson Liu$^{1}$\footnotemark[1] \quad
  Rui Song$^{1}$\footnotemark[1] \quad
  Tony (Xuewei) Qi$^{2}$\thanks{Project lead. Corresponding author. Contact: \texttt{qixuewei@gmail.com}}  \quad
  \textbf{Yiluan Guo}$^{2}$ \\
  \textbf{Zewei Zhou}$^{1,2}$ \;
  \textbf{Hanyin Zhang}$^{1}$ \;
  \textbf{Mingxuan Gao}$^{1}$ \; 
  \textbf{Valeria Xu}$^{1}$ \;  
  \textbf{Jiali Chen}$^{1}$ \; \\
   \textbf{Yun Zhang}$^{1}$ \;
  \textbf{Tianhui Cai}$^{1}$ \;
  \textbf{Ruining Yang}$^{2}$  \; 
  \textbf{Yishan Shen}$^{2}$ \;
  \textbf{Jiaqi Ma}$^{1}$ \\[0.1cm]
  $^{1}$University of California, Los Angeles \quad
  $^{2}$Motional  \\ [0.1cm]
 \small \tt{\href{https://nureasoning.github.io/}{https://nureasoning.github.io/}}
}

\begin{document}

\maketitle

\begin{abstract}
Reasoning is essential for autonomous driving (AD) in long-tail scenarios, where vehicles must apply commonsense knowledge, understand spatial relations, infer agent interactions, and make safe decisions. However, existing AD datasets and benchmarks mainly target perception, prediction, or planning, and provide limited supervision for reasoning over realistic long-tail driving scenes. We introduce \textbf{nuReasoning}, a large-scale real-world dataset and benchmark for reasoning-centric AD. Following the lineage of nuScenes and nuPlan, nuReasoning advances real-world AD datasets and benchmarks toward reasoning in long-tail driving scenarios. The dataset contains 20,000 clips, each 20 seconds long, collected across multiple cities, with synchronized multi-camera images, LiDAR data, HD maps, object annotations, and human-verified reasoning annotations spanning \emph{Spatial Reasoning}, \emph{Decision Reasoning}, and \emph{Counterfactual Reasoning}. Unlike prior datasets that focus primarily on visual question answering, nuReasoning supports both reasoning evaluation and planning evaluation, enabling a direct study of how reasoning supervision affects driving performance. Experiments show that fine-tuning VLMs on nuReasoning substantially improves driving-specific question answering, while incorporating reasoning supervision into VLA training improves planning performance even when textual reasoning outputs are disabled at inference time. These results establish nuReasoning as a foundation for evaluating and improving robust, interpretable, reasoning-driven AD systems in realistic long-tail settings.
\end{abstract}

\section{Introduction}
Autonomous driving (AD) has advanced rapidly over the past decade, driven by progress in perception, prediction, and planning \cite{hu2022st, li2024bevformer, song2024collaborative, zhou2024v2xpnp, liaomaptr, huang2023gameformer, cai2025relmap, gross2025ipformer, Dauner2023CORL, huang2025mdg, huang2025gen}. Recently, end-to-end methods have emerged as a dominant paradigm by directly mapping sensor inputs to driving actions within a unified framework \cite{chen2024end, jiang2023vad, zheng2024genad, wu2022trajectory, hu2023planning, liao2025diffusiondrive, jiadrivetransformer, zhang2025future, zhang2026perception, tian2025simscale}. However, most end-to-end systems are still primarily trained through expert trajectory imitation, which can limit their ability to generalize to rare, unseen, and long-tail scenarios that require deeper scene understanding and reasoning.
Generative AI provides a new opportunity to build more capable and interpretable AD systems \cite{wang2025generative}. Vision-language models (VLMs) have shown strong potential for perception, semantic understanding, and visual reasoning \cite{cai2026driving, xie2025vlms, hwangemma, sima2024drivelm, wang2025omnidrive, li2024driving, xie2025s4, tan2025latent, nie2024reason2drive}, while vision-language-action (VLA) models extend these capabilities to decision-making and planning \cite{hu2025vision, zhou2025autovla, zhou2026opendrivevla, wang2025alpamayo, zengfuturesightdrive, peng2025colavla, liu2026uni, fu2025minddrive, lu2025real, shao2024lmdrive}. A central promise of these models is \textbf{reasoning}: inferring spatial relationships, interpreting agent intent, understanding scene context, anticipating consequences, and making informed decisions in complex environments. However, reasoning remains underdeveloped in current AD systems, largely because existing datasets and benchmarks provide limited supervision and evaluation for reasoning in realistic driving scenarios.

\begin{figure}[t]
    \centering
    \includegraphics[width=\linewidth]{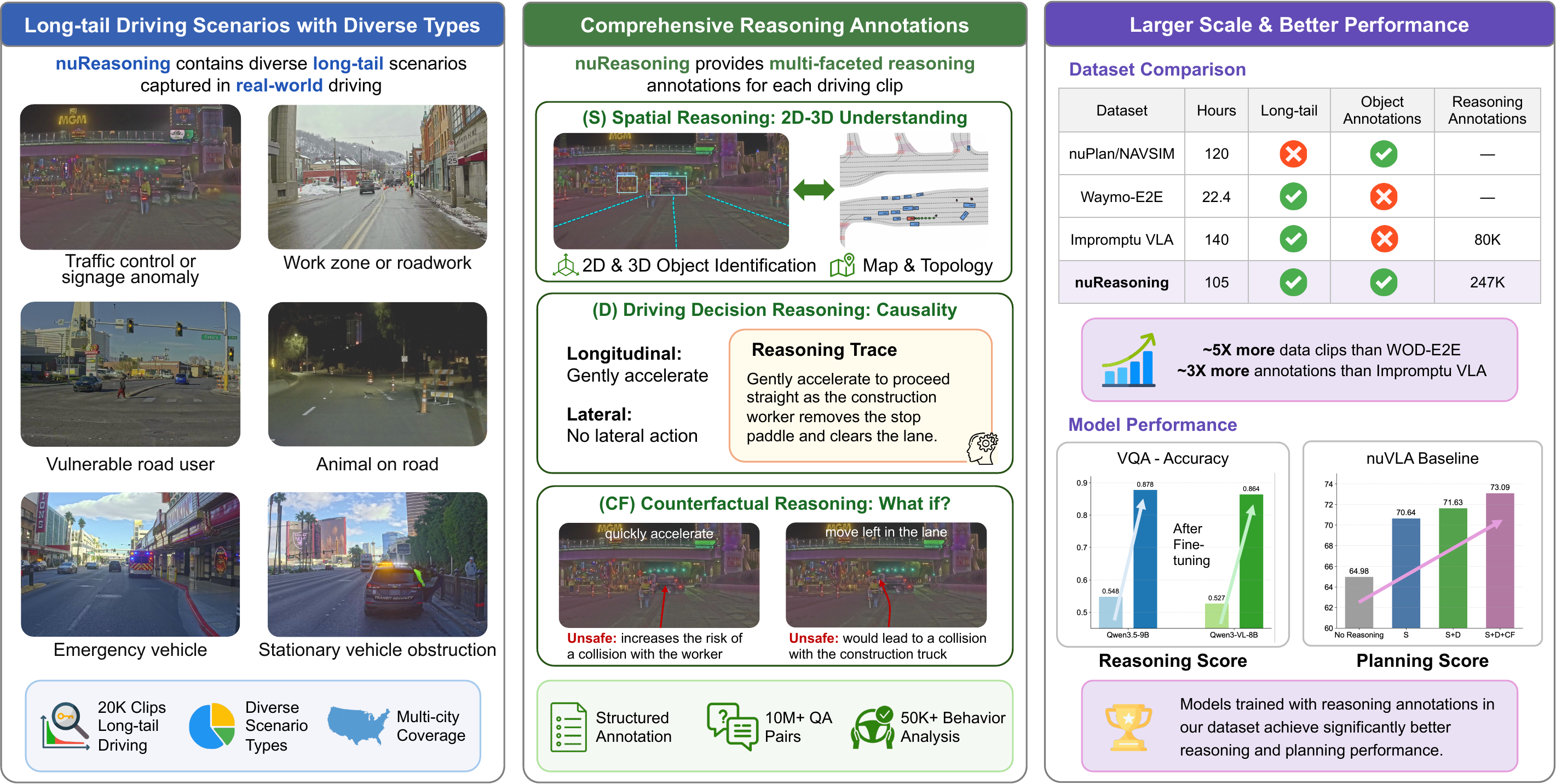}
    \caption{\textbf{nuReasoning} is a large-scale real-world long-tail driving dataset containing 20K 20-second clips across diverse scenario types. The dataset provides high-quality reasoning annotations spanning spatial reasoning, driving decisions, and counterfactual reasoning. Compared with prior datasets, nuReasoning offers substantially larger-scale long-tail driving data and richer reasoning annotations, enabling models trained on it to achieve significantly improved reasoning and planning performance.}
    \label{dataset}
    \vspace{-0.4cm}
\end{figure}

Existing AD datasets and benchmarks mainly target perception \cite{caesar2020nuscenes, sun2020scalability}, trajectory prediction \cite{ettinger2021large, Chang_2019_CVPR, wang2026trends}, or planning \cite{caesar2021nuplan, xu2025wod,zhao2026bridgesim,yu2025drivee2e}. While these datasets have been essential for AD research, they provide limited support for reasoning-centric learning. Recent datasets introduce VLM-oriented reasoning tasks \cite{xie2025vlms, park2025nuplanqa, qian2024nuscenes, marcu2024lingoqa, gu2026accelerating, yu2025waymoqa}, but typically focus on question answering and do not evaluate downstream planning. VLA datasets \cite{arai2025covla, chi2025impromptu} are often built from routine driving data and provide limited coverage of long-tail, reasoning-intensive cases. Moreover, existing annotations are usually narrow in scope, emphasizing scene descriptions while lacking spatial-temporal, causal, and counterfactual reasoning. Consequently, current VLA systems remain brittle in long-tail scenarios and struggle with multi-view, temporal, and causal understanding \cite{xie2025s4, wei2025spatial, li2025spacedrive, liu2025reasonplan, peng2025counterfactual}.

To address these limitations, we introduce \textbf{nuReasoning}, a large-scale real-world dataset and benchmark for reasoning and end-to-end learning in AD. Following the dataset and benchmark lineage of \emph{nuScenes} \cite{caesar2020nuscenes} and \emph{nuPlan} \cite{caesar2021nuplan, karnchanachari2024towards}, nuReasoning extends this ecosystem from perception and planning toward reasoning-centric autonomous driving. The dataset contains 20K real-world driving clips, each 20 seconds long at 10~Hz, mined from large-scale driving logs to maximize scenario diversity and difficulty. Each clip includes synchronized multi-modal inputs, including multi-camera images, LiDAR data, HD maps, and object annotations, together with three complementary reasoning annotations: \emph{Spatial Reasoning}, \emph{Decision Reasoning}, and \emph{Counterfactual Reasoning}. These annotations are produced by automated annotation pipelines and validated by human experts to ensure scalability and quality. Based on this dataset, the \textbf{nuReasoning Benchmark} evaluates both reasoning accuracy and planning performance, providing a unified benchmark for VLM and VLA systems. The key features of nuReasoning and its distinctions from prior datasets are illustrated in \cref{dataset}.

Experimental results show that nuReasoning provides consistent benefits across both reasoning and planning. Fine-tuning VLMs on nuReasoning substantially improves driving-specific reasoning (visual question answering), especially for spatial grounding, decision-making, and counterfactual assessment. Incorporating reasoning supervision into VLA training further improves planning performance, indicating that reasoning annotations can serve as useful intermediate supervision for learning better planning representations. The main contributions of this paper are summarized as:
\begin{enumerate}[leftmargin=1.5em, itemsep=0.5em, topsep=0.5em, parsep=0pt, partopsep=0pt]
\item We introduce \textbf{nuReasoning}, to the best of our knowledge, the first large-scale real-world dataset focused on long-tail driving scenarios with comprehensive reasoning annotations spanning spatial, decision, and counterfactual reasoning.
\item We propose the \textbf{nuReasoning Benchmark}, which evaluates both reasoning and planning performance, providing a unified testbed for VLMs, VLA models, and end-to-end driving models.
\item We benchmark several representative VLMs and planning models, and introduce \textbf{nuVLA}, a strong VLA baseline for autonomous driving. Results show that reasoning supervision from nuReasoning consistently improves both VLM reasoning ability and VLA planning performance.
\end{enumerate}

\section{Related Work}

\textbf{End-to-End AD.}
End-to-end driving models replace modular pipelines with unified sensor-to-action learning. Representative methods include UniAD \cite{hu2023planning}, VAD \cite{jiang2023vad}, ParaDrive \cite{weng2024drive}, SparseDrive \cite{sun2025sparsedrive}, DiffusionDrive \cite{liao2025diffusiondrive}, and BevAD \cite{holtz2026matters}. Although these methods have substantially advanced end-to-end perception, prediction, and planning, their supervision is still largely dominated by trajectories and detections, limiting their ability to learn explicit reasoning in ambiguous and long-tail scenarios.

\textbf{Vision-Language-Action Models for AD.}
Building on the end-to-end driving paradigm, recent VLA models incorporate language into scene understanding, decision-making, and planning. Representative works include DriveVLM \cite{tian2024drivevlm}, SimLingo \cite{renz2025simlingo}, Orion \cite{fu2025orion}, AutoVLA \cite{zhou2025autovla}, OpenDriveVLA \cite{zhou2026opendrivevla}, Alpamayo \cite{wang2025alpamayo}, and Reasoning-VLA \cite{zhang2025reasoning}. More recent reasoning-oriented methods highlight the importance of reasoning: S4-Driver \cite{xie2025s4} emphasizes spatial-temporal reasoning, SpaceDrive \cite{li2025spacedrive} improves spatial awareness, Counterfactual VLA \cite{peng2025counterfactual} introduces self-reflective counterfactual reasoning, and OmniDrive \cite{wang2025omnidrive} bridges planning and language-based counterfactual reasoning. While these works demonstrate the promise of reasoning-centric driving, progress remains constrained by the lack of datasets and benchmarks that systematically support and evaluate diverse reasoning abilities. nuReasoning addresses this gap by providing supervision and evaluation for spatial-temporal, decision, and counterfactual reasoning, and measuring their impact on planning.

\textbf{AD Datasets and Benchmarks.}
Large-scale datasets such as nuScenes \cite{caesar2020nuscenes}, Waymo Open Motion Dataset \cite{ettinger2021large}, and Argoverse \cite{wilson2023argoverse} have driven progress in perception, forecasting, and simulation. For planning, nuPlan provides a large-scale real-world benchmark \cite{caesar2021nuplan, karnchanachari2024towards}, while NAVSIM focuses on end-to-end driving evaluation \cite{dauner2024navsim, Cao2025CORL}. WOD-E2E \cite{xu2025wod} extends the data landscape toward long-tail end-to-end driving but lacks reasoning supervision. For closed-loop driving evaluation, CARLA-based benchmarks such as Bench2Drive provide standardized testbeds \cite{dosovitskiy2017carla, jia2024bench2drive, gerstenecker2026fail2drive}. In contrast, nuReasoning targets long-tail scenarios at scale and introduces structured reasoning annotations with a comprehensive reasoning and planning benchmark.

\textbf{Reasoning Annotations and Benchmarks.}
Recent work introduces language supervision through question answering, instruction following, and VLA-oriented annotation. DriveLM \cite{sima2024drivelm}, MetaVQA \cite{wang2025embodied}, LingoQA \cite{marcu2024lingoqa}, NuScenes-QA \cite{qian2024nuscenes}, WaymoQA \cite{yu2025waymoqa}, and nuPlan-QA \cite{park2025nuplanqa} cast driving understanding as QA, while nuInstruct \cite{ding2024holistic} extends this paradigm with multimodal instruction-following. Benchmarks such as DriveBench \cite{xie2025vlms}, DVBench \cite{zeng2025vision}, and Bench2Drive-VL \cite{jia2026bench2drive} evaluate VLM reasoning performance. CoVLA \cite{arai2025covla} provides large-scale trajectory-language pairs, and Impromptu VLA \cite{chi2025impromptu} focuses on planning-oriented QA and action supervision. While these datasets advance language-grounded driving, they do not systematically cover diverse reasoning formats in long-tail scenarios. In contrast, nuReasoning provides structured supervision for spatial-temporal, decision, and counterfactual reasoning, together with evaluation of their impact on downstream planning.

\section{nuReasoning Dataset}

\subsection{Data Structure}
nuReasoning adopts a clip-based data structure: each clip spans 20 seconds at 10~Hz and is centered on a decision-critical keyframe, with 10 seconds of preceding context and 10 seconds of subsequent data. In particular, we use the clip as the basic unit for storage, indexing, and annotation. Each clip contains synchronized multi-view camera images, LiDAR point clouds, sensor calibrations, ego-vehicle states, 3D bounding-box annotations of surrounding objects, traffic-light states, a local static vectorized map, the ego route, and a high-level navigation command. In addition, each clip carries human-verified spatial, decision, and counterfactual reasoning annotations at lower frequencies (1~Hz for spatial and 0.2~Hz for decision and counterfactual), capturing scene understanding and decision-making throughout the clip. 

\begin{figure}[t]
    \centering
    \includegraphics[width=\linewidth]{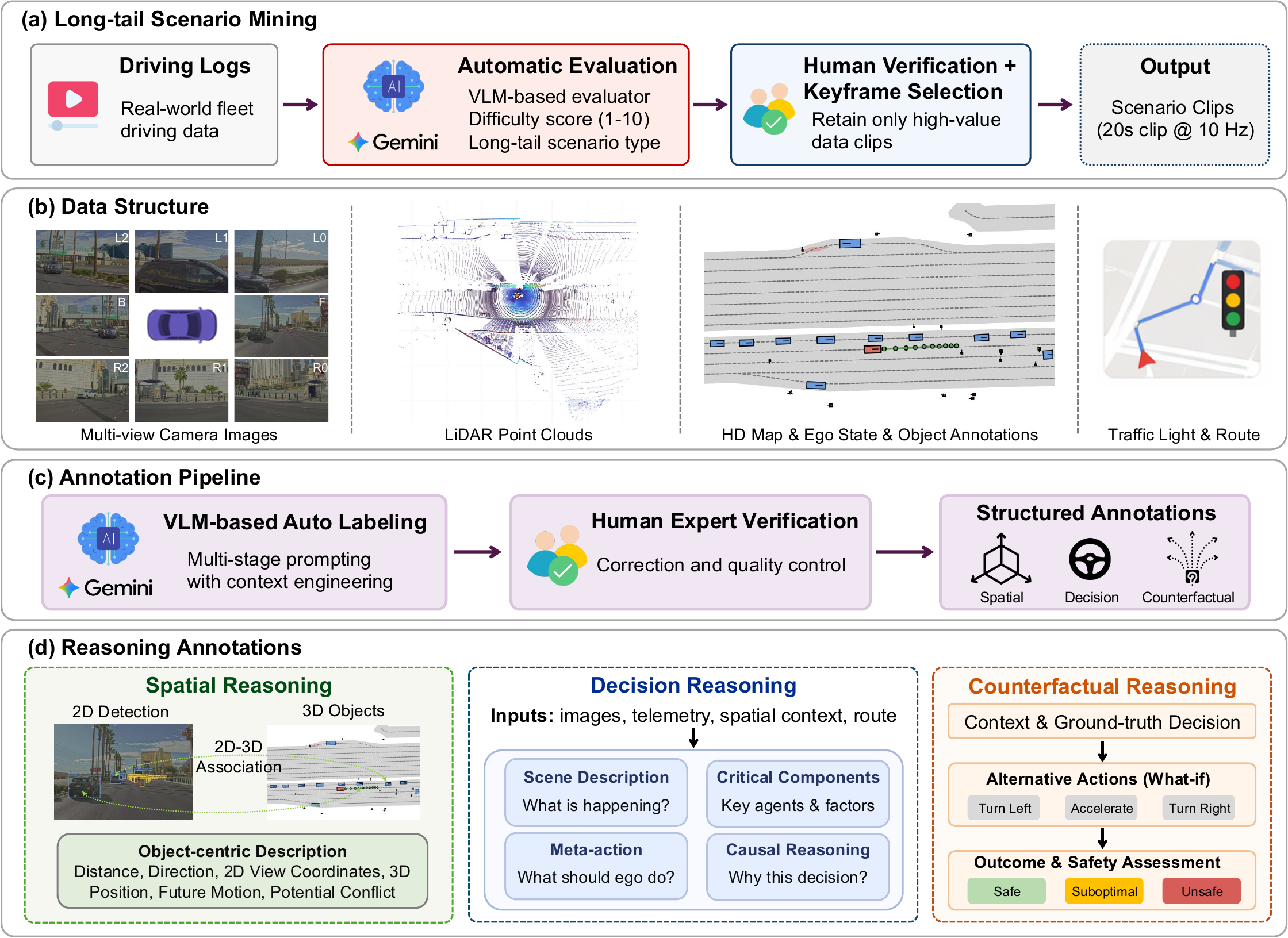}
    \caption{Overview of the long-tail data mining and annotation pipeline. (a) Internal fleet driving logs are segmented and scored by a VLM-based evaluator for scenario type and difficulty, followed by human verification and keyframe selection. (b) Each selected 20-second clip (10~Hz) contains multi-view camera images, LiDAR point clouds, HD maps, ego states, object annotations, traffic lights, and route paths. (c) Annotation combines VLM-based auto-labeling with human verification and correction to ensure quality. (d) Reasoning annotations span spatial, decision-making, and counterfactual dimensions, supporting diverse downstream tasks and reasoning capabilities.}
    \label{fig:pipeline}
    \vspace{-0.4cm}
\end{figure}

The reasoning annotations fall into three categories: \emph{Spatial}, \emph{Decision}, and \emph{Counterfactual}. Spatial reasoning describes the scene and surrounding agents using both per-camera image observations and 3D geometry in the ego-centric coordinate system, including relations such as distance, motion state, and potential conflicts. Decision reasoning explains the ego vehicle's intended behavior as a structured meta-action, together with the scene context that motivates it. Counterfactual reasoning captures the consequences of plausible alternatives to the ground-truth action, labeled as safe, suboptimal, or unsafe. In total, the dataset contains 20K clips, divided into 17K for training, 2K for validation, and 1K reserved as a private test split for evaluation. The data processing pipeline, annotation procedure, and dataset structure are illustrated in \cref{fig:pipeline}.

\subsection{Data Selection and Processing}
We mine Motional's engineering-prefiltered AV driving logs and segment them into 30-second video clips. A VLM-based evaluator built on Gemini 3.1 Pro assesses each video clip using front-camera observations. The evaluator assigns a driving difficulty score from 1 to 10, where 1 denotes nominal driving and 10 denotes highly challenging or rare scenarios, and categorizes the clip into predefined long-tail scenario types, such as vulnerable road users, work zones or roadwork, and perception degradation. Selection follows a \textbf{behavior-centric criterion}: unusual agents, road conditions, or events must induce a meaningful change in ego-vehicle behavior, such as braking, yielding, stopping, or lane changing. The mere presence of unusual elements is insufficient for inclusion. Video clips scoring at or above 5 are passed to human experts, who verify their scenario types and difficulty levels. 81.72\% of them are confirmed as valid long-tail scenarios, and the rest are excluded from the final dataset.

Human experts then identify the decision-critical keyframe, i.e., the moment at which the ego vehicle must make a behavioral decision, following~\cite{wang2025alpamayo}. For each keyframe, we extract 10 seconds of preceding and 10 seconds of subsequent data from the original driving log to form a 20-second data clip. All sensor modalities and annotations are synchronized at 10~Hz, and each data clip is paired with a local static vectorized map. Further details are provided in Appendix~\cref{eval}.

\begin{figure}[t]
    \centering
    \includegraphics[width=\linewidth]{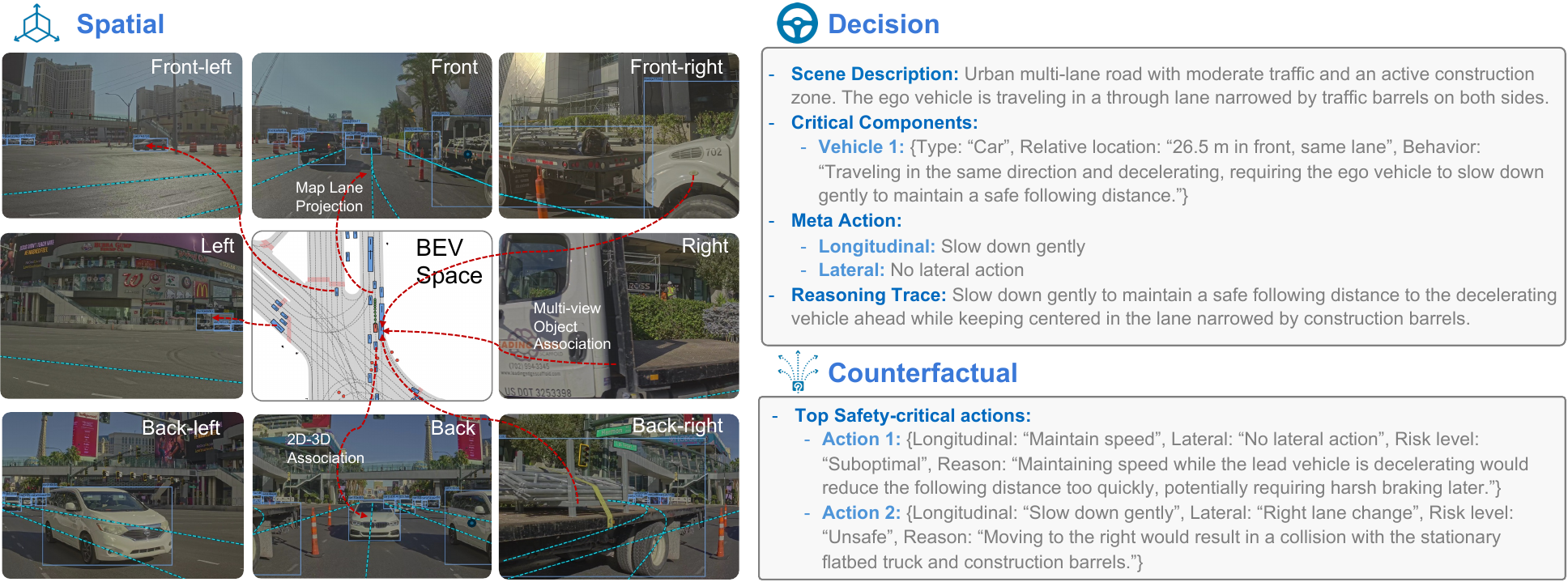}
    \caption{Example of reasoning annotation in the nuReasoning dataset. The frame is annotated with \textit{Spatial}, \textit{Decision}, and \textit{Counterfactual} reasoning to capture scene understanding, meta-actions, and analysis of alternative actions and their risks and outcomes.}
    \label{fig:annotation}
    \vspace{-0.4cm}
\end{figure}

\subsection{Reasoning Annotation}
\label{annotation}
We annotate each clip with three reasoning types: spatial reasoning at 1~Hz starting 3 seconds into the clip, and decision and counterfactual reasoning at 0.2~Hz starting 5 seconds before the keyframe. \cref{fig:annotation} illustrates representative annotations, with the detailed pipeline provided in Appendix~\cref{app:annotation}.

\textbf{Spatial Reasoning.}
We use Gemini 3 Flash to identify driving-relevant 2D objects in multi-view images, while 3D bounding boxes, track identities, positions, and velocities are obtained from an internal offline perception pipeline using multi-frame point clouds. The 2D detections are associated with 3D tracks through calibrated geometric projection and overlap-based matching. Spatial attributes, including relative positions, distances, velocities, and map relationships, are then computed from the associated 3D tracks and map geometry. Only reliably matched objects receive metric 3D attributes, while unmatched or ambiguous detections remain unassociated.

\textbf{Decision Reasoning.}
We use Gemini 3.1 Pro with temporal multi-view observations, ego telemetry, spatial and map context, route commands, and ground-truth future trajectories to generate four components: \emph{Scene Description}, \emph{Critical Components}, \emph{Driving Decision}, and \emph{Reasoning Trace}. The future trajectory grounds the longitudinal and lateral meta-actions in the demonstrated behavior, while critical scene components are identified from past and current observations. These annotations provide \emph{future-grounded explanations} of observed decisions rather than uniquely optimal actions or recovered driver intentions.

\textbf{Counterfactual Reasoning.}
Using the demonstrated decision as a reference, we evaluate plausible alternative longitudinal and lateral actions under the same scene context. Each alternative is assigned a risk level: \emph{Safe} for feasible actions without safety or regulatory concerns, \emph{Suboptimal} for actions that unnecessarily compromise safety margins, comfort, efficiency, or progress, and \emph{Unsafe} for actions posing concrete collision, traffic-rule violation, or drivable-area risks. These labels represent context-dependent risk assessments rather than observed outcomes of executed alternatives.

\textbf{Human Verification.}
Trained annotators verify the automatically generated annotations through a structured quality-control process. Decision and counterfactual annotations undergo two-stage expert review, where one expert checks the initial labels and another independently audits the results. Incorrect annotations are corrected, while unresolved ambiguous cases are excluded. As summarized in \cref{tab:annotation_quality}, pre-correction agreement reaches 81.72\% for scenario selection, 84.69\% for decision reasoning, and 76.11\% for counterfactual reasoning. 

\begin{table}[t]
\centering
\caption{Annotation quality control in nuReasoning. Reported rates reflect pre-correction VLM-human agreement (or scenario confirmation), rather than final-label accuracy. All retained annotations undergo quality verification.}
\label{tab:annotation_quality}
\footnotesize
\setlength{\tabcolsep}{4pt}
\renewcommand{\arraystretch}{1.1}
\begin{tabular}{p{0.23\linewidth}c|p{0.58\linewidth}}
\toprule
Annotation & Agreement & Quality control measures \\
\midrule
Scenario type/difficulty
& 81.72\%
& Experts verify scenario types, difficulty, and behavioral relevance; select critical keyframes and exclude invalid clips. \\

Spatial reasoning
& --
& Validate 2D-3D associations through calibrated projection and geometric consistency; exclude uncertain matches. \\

Decision reasoning
& 84.69\%
& Two-stage expert review verifies meta-actions, critical scene components, and reasoning traces against observed behavior and scene context. \\

Counterfactual reasoning
& 76.11\%
& Two-stage expert review checks alternative actions, risk levels, and explanations against scene geometry, traffic rules, and agent interactions. \\
\bottomrule
\end{tabular}
\vspace{-0.3cm}
\end{table}

\textbf{VQA Data Generation.}
We convert the verified structured annotations into VQA pairs grounded in multi-view images and reasoning labels. The questions cover perception, spatial relationships, motion prediction, map understanding, decision-making, and counterfactual risk assessment, with multiple-choice, text-answer, and numerical formats. This process produces more than 10M QA pairs in total. To balance coverage and training efficiency, we subsample 167K QA pairs for training.

\subsection{Dataset Statistics}
We summarize nuReasoning and compare it with representative AD and reasoning datasets in \cref{data_stats}. nuReasoning contains 20K clips, totaling approximately 105 hours of curated long-tail driving data collected across multiple U.S. cities, including Las Vegas, Pittsburgh, Los Angeles, and Boston. Each clip is centered on a decision-critical keyframe and includes synchronized multi-modal inputs, object annotations, and a local map. Here, a ``frame'' denotes a reasoning-annotated timestep: across the 19K annotated clips, nuReasoning provides 247K spatial frames sampled at 1~Hz and 57K decision/counterfactual frames sampled at 0.2~Hz.
Compared with prior datasets, nuReasoning uniquely combines long-tail coverage, object and map annotations, and reasoning supervision spanning spatial, decision, and counterfactual reasoning, enabling the evaluation of both reasoning ability and downstream planning.

\newcommand{\cmark}{\textcolor{green!60!black}{\checkmark}}
\newcommand{\xmark}{\textcolor{red!70!black}{$\times$}}
\newcommand{\pmark}{\textcolor{orange!80!black}{Partial}}

\begin{table}[t]
  \caption{Comparison of {nuReasoning} with representative autonomous driving and reasoning datasets. \textit{``Long-tail''} indicates explicit coverage of rare or challenging scenarios. For reasoning datasets, \# Frames refers to annotated frames or timesteps used for reasoning supervision. \textit{Spa}, \textit{Dec}, and \textit{CF} denote spatial, decision, and counterfactual reasoning.}
  \label{data_stats}
  \centering
  \renewcommand{\arraystretch}{1.1}
  \setlength{\tabcolsep}{4pt}
  \footnotesize
  \begin{tabular}{l|cccccc}
    \toprule
    Dataset & Hours & Long-tail & Obj./Map Ann. & Reasoning Ann. & \# Frames & Type \\
    \midrule
    nuScenes \cite{caesar2020nuscenes} & 5.6 & \xmark & Obj. + Map & \xmark & -- & --\\
    Argoverse \cite{wilson2023argoverse} & 4.2 & \xmark & Obj. + Map & \xmark & -- & --\\
    WOD \cite{sun2020scalability} & 6.4 & \xmark & Obj. + Map & \xmark & -- & --\\
    nuPlan (NAVSIM) \cite{caesar2021nuplan} & 120 & \xmark & Obj. + Map & \xmark & -- & --\\
    WOD-E2E \cite{xu2025wod} & 22.3 & \cmark & \xmark & \xmark & -- & --\\
    \midrule
    DriveLM-nuScenes \cite{sima2024drivelm} & 5.6 & \xmark & Obj. + Map & \cmark & 4,871 & Spa + Dec\\
    nuPlan-QA \cite{park2025nuplanqa} & 120 & \xmark & Obj. + Map & \cmark & 90K & Spa + Dec\\
    CoVLA \cite{arai2025covla} & 83 & \xmark & \xmark & \cmark & 100K & Dec \\
    Physical AI AV Dataset \cite{wang2025alpamayo} & 1700 & \xmark & Obj. & \cmark & 2,077$^\dagger$ & Dec \\
    ImpromptuVLA \cite{chi2025impromptu} & 144 & \cmark & \xmark & \cmark & 80K & Dec \\
    \midrule
    \rowcolor{blue!10}
    \textbf{nuReasoning (ours)} & 105 & \cmark & Obj. + Map  & \cmark & \textbf{247K} & \textbf{Spa + Dec + CF} \\
    \bottomrule
  \end{tabular}
  \begin{flushleft}
  \footnotesize $^\dagger$Denotes reasoning labels at distinct annotated timesteps from the publicly available dataset as of October 2026.
  \end{flushleft}
  \vspace{-0.4cm}
\end{table}

\section{nuReasoning Benchmark}

\subsection{Reasoning Evaluation}
\label{sec:reasoning_evaluation}
\begin{figure}[t]
    \centering
    \includegraphics[width=\linewidth]{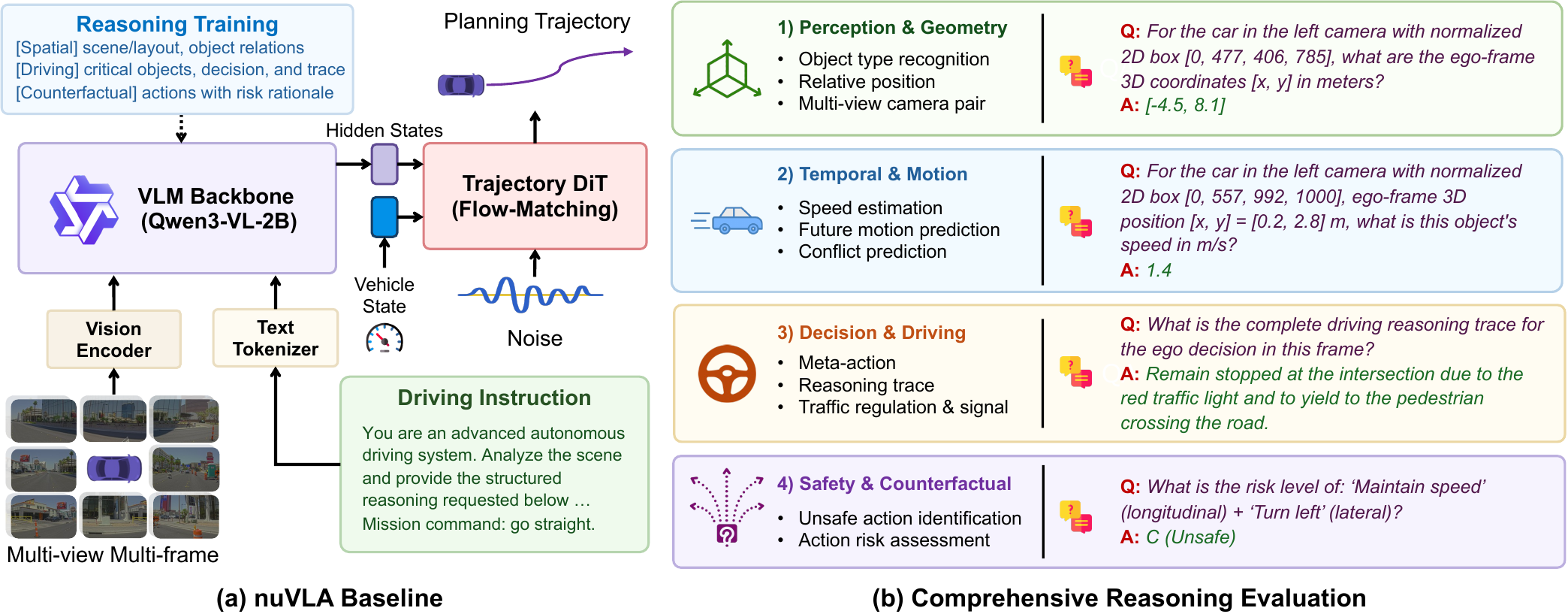}
    \caption{Overview of the nuVLA baseline and the reasoning evaluation benchmark. (a) nuVLA takes multi-view, multi-frame camera images and driving instructions as input. The encoded hidden states are fed into a trajectory DiT to generate future planning trajectories. The VLM backbone is trained with different types of reasoning supervision. (b) For reasoning evaluation, we assess diverse capabilities, including 3D geometric understanding, future object motion estimation, decision-making, and counterfactual reasoning.}
    \label{models}
    \vspace{-0.4cm}
\end{figure}

We introduce the reasoning benchmark to evaluate whether VLMs can reason about driving scenes beyond generic visual recognition. Inspired by prior driving VLM benchmarks \cite{sima2024drivelm,park2025nuplanqa,xie2025vlms}, we group question types by the driving capabilities they assess and report results across four categories: \emph{Geometry}, \emph{Motion}, \emph{Driving}, and \emph{Counterfactual}, as illustrated in \cref{models}(b).

\textbf{Evaluation Metrics.}
Each question is evaluated with the metric matching its answer format. We report choice accuracy for multiple-choice questions, tolerance-based accuracy for scalar answers, and hit rate and L2 error for coordinates. For future trajectories, we report the mean waypoint L2 error and the hit rate, where a hit requires the mean waypoint error over the 5-second horizon to be within 1~m. Token-level F1 and ROUGE-L serve only as auxiliary measures for free-form reasoning traces. Relying mainly on structured metrics reduces the dependence on language similarity or LLM-based judging \cite{sima2024drivelm,marcu2024lingoqa,xie2025vlms,wang2025omnidrive}.

\textbf{Evaluation Settings.}
All models are evaluated on the same held-out test set, which contains 16,437 evaluable questions, and receive the same multi-view, multi-frame camera inputs and textual context. We parse model outputs deterministically: choice answers are normalized to option labels or option text, scalar answers are parsed as numerical values, and coordinate or trajectory answers are parsed into structured numeric outputs. Unparseable outputs are counted as incorrect for tolerance-based and hit-rate metrics. Detailed settings and the full subcategory mapping are provided in Appendix~\cref{app:eval_protocol}.

\textbf{VLM Baselines.}
We report both proprietary and open-weight VLM baselines. The proprietary baselines are Gemini-3.1-Pro~\cite{google2026gemini31procard} and Gemini-3-Flash~\cite{google2026geminiflash}. The open-weight base models are Qwen3.5-4B~\cite{team2026qwen3}, Qwen3.5-9B~\cite{team2026qwen3}, and Qwen3-VL-8B-Instruct~\cite{bai2025qwen3}. We fine-tune these open-weight models on the nuReasoning training set and compare each fine-tuned model against its base model. In addition, we include NVIDIA Cosmos-Reason2-8B~\cite{nvidia2025cosmosreason2} as a strong external baseline for zero-shot evaluation.

\subsection{Planning Evaluation}
We evaluate the 5-second ego trajectory predicted at each keyframe against the recorded future trajectory, the future states of surrounding objects, and the HD map. The benchmark follows an \emph{open-loop} protocol in which surrounding agents replay their recorded trajectories.

\textbf{Evaluation Metrics.}
The nuReasoning planning score (NPS) combines five normalized sub-scores: non-at-fault collision (NC), drivable-area compliance (DA), ego route progress (EP), comfort (Comf), and human-likeness (HL):
\begin{equation}
\mathrm{NPS}=s_{\mathrm{NC}}\,s_{\mathrm{DA}}
\left(0.3\,s_{\mathrm{EP}}+0.2\,s_{\mathrm{Comf}}+0.5\,s_{\mathrm{HL}}\right).
\label{eq:nps}
\end{equation}
NC and DA act as multiplicative safety gates, so unsafe trajectories are penalized before progress, comfort, and similarity to the expert are considered. EP rewards progress along the intended route, Comf encourages smooth motion, and HL measures consistency with the recorded expert trajectory. We assign HL the largest weight because driving behavior in long-tail scenarios is highly context-dependent and the recorded expert trajectory is the most informative reference available. Scores are computed per clip, averaged over the test set, and reported on a 0-100 scale. We additionally report the individual sub-scores and the average displacement error (ADE).

\textbf{Planning Baselines.}
Our nuVLA baseline combines a Qwen3-VL-2B backbone~\cite{bai2025qwen3} with a flow-matching trajectory DiT~\cite{bjorck2025gr00t}. The DiT cross-attends to the prompt-side VLM features and predicts a 5-second ego-frame trajectory (\cref{models}(a); Appendix~\cref{app:vla}). We compare it with the autoregressive VLA AutoVLA~\cite{zhou2025autovla}, the zero-shot Alpamayo-1.5~\cite{wang2025alpamayo}, and two end-to-end planners, DiffusionDrive~\cite{liao2025diffusiondrive} and the BEV-based UniAD~\cite{hu2023planning}. All baselines except Alpamayo-1.5 are trained on the nuReasoning training split. The nuVLA variants share the same architecture and training protocol and differ only in the type of reasoning supervision, which isolates the contribution of each type.

\section{Experiments}

\subsection{Implementation Details}
All trainable models are trained on 8 NVIDIA A100 GPUs. For nuVLA, each sample uses eight surrounding camera views at the current timestep and one second earlier, with images resized to $448\times448$. VLM/VLA fine-tuning uses a per-GPU batch size of 2 and four gradient-accumulation steps, yielding an effective batch size of 64, with Low-Rank Adaptation (LoRA) of the vision-language backbone. nuVLA uses AdamW with initial learning rates of $1\times10^{-4}$ for the action head and $5\times10^{-5}$ for the backbone, linear warm-up, and cosine annealing. All nuVLA variants use the same reasoning-annotated clip pool, including the planning-only control. Appendix~\cref{app:vla} provides more architecture and training details.

\begin{table}[t]
\centering
\setlength{\tabcolsep}{4.5pt}
\renewcommand{\arraystretch}{1.15}
\caption{Reasoning results under the four main capabilities, grouped by category; each
metric follows its answer format, with multi-frame input when available. FT denotes
fine-tuning. Ch.\ is choice accuracy, Num.\ is numerical accuracy within tolerance,
Coord.\ Hit is coordinate hit rate, Traj.\ Hit is Hit@1m, and Text F1 is token-level F1
(all in \%). Coord.\ L2 averages ego-frame position errors (m) and image-coordinate
errors (0-1000 scale), so it compares models but is not a physical distance; Traj.\ L2
is the mean waypoint error in meters. ``--'' denotes an unparsable trajectory output.}
\label{tab:reasoning_four_main_category_results}
\scriptsize
\setlength{\tabcolsep}{1.3pt}
\begin{tabular}{l|c|cccc|cccc|cc|c}
\toprule
\multirow{2}{*}{Model} & \multirow{2}{*}{FT} & \multicolumn{4}{c|}{Geometry} & \multicolumn{4}{c|}{Motion} & \multicolumn{2}{c|}{Driving} & Counterfactual \\
\cmidrule(lr){3-6}\cmidrule(lr){7-10}\cmidrule(lr){11-12}\cmidrule(lr){13-13}
& & Ch. $\uparrow$ & Num. $\uparrow$ & Coord. Hit $\uparrow$ & Coord. L2 $\downarrow$ & Ch. $\uparrow$ & Num. $\uparrow$ & Traj. Hit $\uparrow$ & Traj. L2 $\downarrow$ & Ch. $\uparrow$ & Text F1 $\uparrow$ & Ch. $\uparrow$ \\
\midrule
Gemini-3.1-Pro & $\times$ & 69.2 & 3.9 & 2.4 & 149.80 & 72.9 & 12.3 & 0.0 & 38.16 & 54.9 & 19.7 & 38.3 \\
Gemini-3-Flash & $\times$ & 55.6 & 3.6 & 2.9 & 169.74 & 72.0 & 11.0 & 0.0 & 46.00 & 52.6 & 20.0 & 56.0 \\
\midrule
Qwen3-VL-8B & $\times$ & 41.2 & 2.8 & 0.1 & 305.04 & 71.8 & 11.7 & 0.0 & 138.76 & 53.6 & 14.1 & 44.7 \\
Cosmos-Reason2-8B & $\times$ & 41.5 & 2.8 & 0.0 & 298.45 & 68.8 & 13.1 & 0.0 & 77.39 & 49.1 & 18.9 & 46.0 \\
\rowcolor{blue!10}
Qwen3-VL-8B & \checkmark & 92.0 & 37.1 & 46.3 & 31.38 & 92.1 & \textbf{25.8} & 0.0 & 19.90 & 70.0 & 41.9 & 81.6 \\
\midrule
Qwen3.5-4B & $\times$ & 38.8 & 2.7 & 0.1 & 451.27 & 48.3 & 5.7 & 0.0 & -- & 42.6 & 20.1 & 47.5 \\
\rowcolor{blue!10}
Qwen3.5-4B & \checkmark & 92.5 & 36.9 & 46.4 & 25.18 & 92.2 & 24.4 & 0.0 & 17.36 & \textbf{78.4} & \textbf{42.9} & 83.3 \\
\midrule
Qwen3.5-9B & $\times$ & 47.9 & 4.6 & 0.3 & 212.17 & 74.9 & 5.3 & 0.0 & 425.68 & 44.9 & 19.5 & 47.2 \\
\rowcolor{blue!10}
Qwen3.5-9B & \checkmark & \textbf{93.3} & \textbf{37.2} & \textbf{52.2} & \textbf{19.53} & \textbf{92.5} & \textbf{25.8} & \textbf{0.2} & \textbf{16.32} & 74.4 & 42.6 & \textbf{83.4} \\
\bottomrule
\end{tabular}
\vspace{-0.4cm}
\end{table}

\subsection{Results on Reasoning Benchmark}
\begin{figure}[t]
    \centering
    \includegraphics[width=\linewidth]{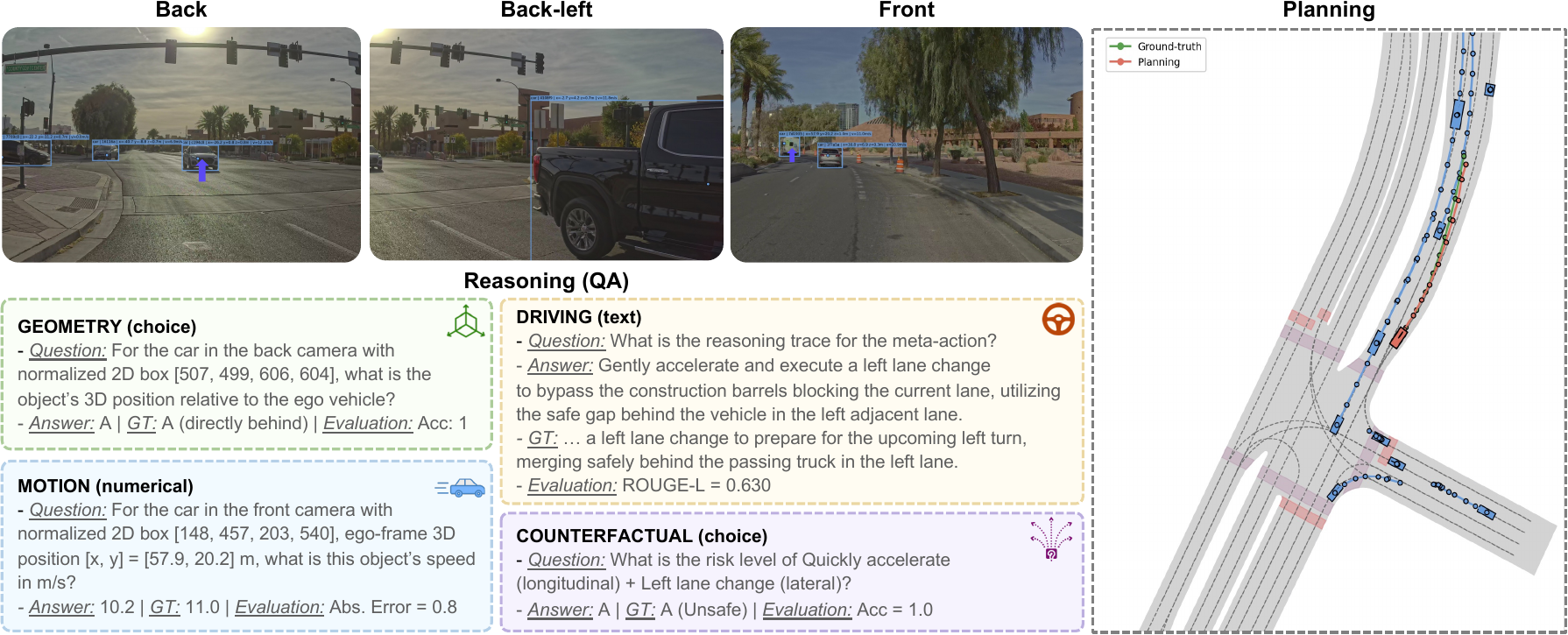}
    \caption{Example of reasoning and planning results on the test set of nuReasoning. Reasoning is evaluated using the fine-tuned Qwen3.5-9B model, while planning is evaluated using the nuVLA model trained with all types of reasoning supervision (S+D+CF).}
    \label{results}
    \vspace{-0.4cm}
\end{figure}

Results in \cref{tab:reasoning_four_main_category_results} reveal a substantial gap between general-purpose VLM capabilities and driving-specific reasoning. Proprietary Gemini models outperform pretrained open-weight models on several choice-based metrics, yet struggle with precise spatial and temporal grounding: coordinate hit rates remain below 3\%, and trajectory Hit@1m is 0\% across all models without task-specific fine-tuning. This suggests that general-purpose pretraining provides semantic understanding but insufficient metric grounding for complex driving scenarios. Fine-tuning on nuReasoning substantially improves all four reasoning capabilities, with open-weight models surpassing proprietary baselines on most metrics. For Qwen3-VL-8B, geometry choice accuracy increases from 41.2\% to 92.0\%, numerical accuracy from 2.8\% to 37.1\%, driving choice accuracy from 53.6\% to 70.0\%, and counterfactual accuracy from 44.7\% to 81.6\%. Similar improvements across Qwen3.5-4B and Qwen3.5-9B demonstrate consistent benefits across model scales. \cref{results} illustrates representative reasoning and planning outputs.

Despite substantial improvements, precise future-motion prediction remains challenging. Fine-tuning considerably reduces trajectory L2 error, but trajectory Hit@1m remains near zero. In contrast, coordinate hit rates reach 46.3\% for Qwen3-VL-8B and 52.2\% for Qwen3.5-9B, indicating greater improvements in spatial grounding than in future-motion prediction. This disparity suggests that predicting temporally evolving trajectories requires capabilities beyond accurate instantaneous scene understanding. Moreover, Cosmos-Reason2-8B~\cite{nvidia2025cosmosreason2}, despite its physical and embodied reasoning pretraining~\cite{azzolini2025cosmosreason1}, achieves limited metric grounding without task-specific fine-tuning. These findings underscore the importance of fine-grained, driving-specific supervision while highlighting future-motion reasoning as a persistent challenge.

\begin{table}[t]
\centering
\setlength{\tabcolsep}{4.5pt}
\renewcommand{\arraystretch}{1.15}
\caption{
Comparison of planning models on the nuReasoning test set. \textit{Planning} denotes trajectory planning supervision. nuVLA uses reasoning supervision with different components: {S} (spatial), {D} (decision), and {CF} (counterfactual).
}
\label{tab:reasoning_planning_comparison}
\scriptsize
\setlength{\tabcolsep}{6pt}
\begin{tabular}{l|l|ccccc|cc}
\toprule
\multirow{2}{*}{Model} & \multirow{2}{*}{Supervision} & \multicolumn{7}{c}{Planning Performance} \\
\cmidrule(lr){3-9}
& & NC $\uparrow$ & DA $\uparrow$ & EP $\uparrow$ & Comf $\uparrow$ & HL $\uparrow$ & NPS $\uparrow$ & ADE $\downarrow$ \\
\midrule
UniAD \cite{hu2023planning}                & Planning + BEV   &  88.87   & 87.62    & 89.62    & 92.62    & 48.80    & 55.65    & 2.054                 \\
DiffusionDrive \cite{liao2025diffusiondrive}      & Planning                          & 90.22   & 88.25    & 90.46   & 94.96    & 51.96   & 57.86    & 1.930  \\
AutoVLA \cite{zhou2025autovla}              & Planning + Decision               & 90.92 & 86.48 & 89.33    & 99.90    & 49.89  & 59.05   & 2.063 \\
Alpamayo-1.5 \cite{wang2025alpamayo}         & Zero-shot                             & 90.26    & 86.13    & 86.51    & 97.93  & 33.79    & 50.45    & 2.925 \\
\midrule
\rowcolor{blue!5}
nuVLA                   & Planning                          & 94.87 & 92.10 & 87.38 & 99.70 & 55.22 & 64.98 & 1.937 \\
\rowcolor{blue!5}
nuVLA (D)               & Planning + Reasoning (D)          & 96.25 & 94.08 & 90.30 & \textbf{99.90} & 61.01 & 70.91 & 1.676 \\
\rowcolor{blue!5}
nuVLA (D + CF)          & Planning + Reasoning (D + CF)     & 96.24    & 95.36    & 90.35    & 99.80    & 61.26    & 72.04   & 1.614 \\
\rowcolor{blue!5}
nuVLA (S)               & Planning + Reasoning (S)          & 95.36 & 92.99 & \textbf{91.52} & 99.61 & 61.65 & 70.64 & 1.597 \\
\rowcolor{blue!5}
nuVLA (S + D)           & Planning + Reasoning (S + D)      & \textbf{96.84} & 94.96    & 90.55    & 99.70    & 60.34    & 71.63   & 1.608 \\
\rowcolor{blue!5}
nuVLA (S + D + CF)      & Planning + Reasoning (S + D + CF) & 96.25 & \textbf{96.25} & 90.73  & 99.70    & \textbf{62.18}   & \textbf{73.09} &  \textbf{1.555}\\
\bottomrule
\end{tabular}
\vspace{-0.3cm}
\label{vla_results}
\end{table}

\subsection{Results on Planning Benchmark}
We train all trainable baselines on nuReasoning and evaluate them over a 5-second planning horizon, while Alpamayo-1.5 is evaluated zero-shot. For nuVLA, ground-truth trajectories supervise the trajectory DiT, while reasoning annotations supervise the VLM backbone; both are jointly optimized. Controlled ablations keep the architecture, training data, planning targets, and optimization fixed, varying only reasoning supervision. Explicit reasoning generation is disabled during inference.

\textbf{Overall planning performance.}
As shown in \cref{vla_results}, nuVLA outperforms competitive driving and VLA baselines. Zero-shot Alpamayo-1.5 achieves 50.45 NPS, highlighting the difficulty of transferring existing models to long-tail scenarios. Full spatial, decision, and counterfactual (S+D+CF) supervision improves NPS from 64.98 to 73.09 and reduces ADE from 1.937 to 1.555~m (19.7\%), demonstrating the benefits of reasoning supervision without test-time reasoning generation.

\textbf{Complementary supervision.}
Different reasoning types provide complementary benefits. Spatial supervision improves route progress, while decision supervision improves collision avoidance and drivable-area compliance. Adding counterfactual supervision to decision reasoning increases NPS from 70.91 to 72.04 by incorporating alternative actions and their risks. However, gains are not strictly additive: adding decision reasoning to spatial supervision reduces human-likeness from 61.65 to 60.34 despite improving NPS, suggesting that high-level decisions do not fully capture precise trajectory execution. Nevertheless, combining all three reasoning types achieves the best overall planning performance.

\subsection{Ablation Studies}
\label{sec:controls_transfer}

\textbf{Cross-dataset transfer.}
We evaluate nuVLA across nuReasoning and NAVSIM without any target-domain fine-tuning. As shown in \cref{tab:controls_transfer}, nuVLA trained on NAVSIM achieves 52.63 NPS on nuReasoning, compared with 64.98 for in-domain planning-only training, a transfer gap of 12.35 points. Conversely, nuVLA trained on nuReasoning with full \emph{S+D+CF} supervision achieves 72.73 PDMS on NAVSIM Navtest, versus 81.95 for the NAVSIM-trained model. These results indicate a substantial distribution shift between the two datasets, while showing that training on long-tail data retains meaningful planning capability on NAVSIM. The gaps may also reflect differences beyond scenario difficulty, such as sensor setup and evaluation protocol, which motivates evaluating on both datasets. Note that NPS and PDMS are comparable only within their respective benchmarks.

\begin{table}[t]
\centering
\caption{Cross-dataset transfer of nuVLA. All models use planning supervision, with optional spatial (S), decision (D), and counterfactual (CF) reasoning supervision. Cross-dataset results are obtained without target-domain fine-tuning. Scores are comparable only within a column.}
\label{tab:controls_transfer}
\small
\setlength{\tabcolsep}{7pt}
\begin{tabular}{llcc}
\toprule
Training data & Reasoning targets & nuReasoning NPS $\uparrow$ & NAVSIM PDMS $\uparrow$ \\
\midrule
NAVSIM & None & 52.63 & 81.95 \\
nuReasoning & None & 64.98 & -- \\
nuReasoning & S+D+CF & \textbf{73.09} & 72.73 \\
\bottomrule
\end{tabular}
\vspace{-0.3cm}
\end{table}

\textbf{Scenario-distribution control.}
Work-zone scenarios make up a large share of the dataset, reflecting their frequency in the driving logs and their diverse configurations, including lane closures, temporary traffic controls, and altered road layouts. Such scenarios call for varied driving decisions and may therefore provide valuable, non-redundant supervision. To examine the effect of this imbalance, we cap work-zone clips at 20\% of the training data while keeping the test set unchanged. As reported in Appendix~\cref{app:workzone_control}, this reduces the NPS of the full model from 73.09 to 68.12 and the coordinate hit rate of Qwen3.5-4B from 46.4\% to 34.3\%, suggesting that work-zone diversity contributes to both reasoning and planning. However, capping also reduces the training volume, so we cannot separate the effect of scenario composition from that of data quantity.

\textbf{Test-time reasoning and score sensitivity.}
We examine whether generating explicit reasoning at inference time improves planning, with detailed results in Appendix~\cref{app:inference_control}. A separately trained variant conditions the trajectory head on generated meta-actions and reasoning traces. It achieves 70.23 NPS at 1711.91~ms per sample, versus 70.91 NPS at 202.10~ms for the standard decision-supervised nuVLA on an NVIDIA RTX Pro 6000. Test-time reasoning thus increases latency by about $8.5\times$ without improving NPS. In addition, the sensitivity analysis in Appendix~\cref{app:metric_sensitivity} shows that full S+D+CF supervision achieves the highest NPS under all five tested weight configurations, and that every reasoning-supervised variant outperforms planning-only training.

\section{Conclusions}
We introduce \textbf{nuReasoning}, a large-scale dataset and benchmark integrating spatial, decision, and counterfactual reasoning with planning in long-tail driving scenarios. Its synchronized multimodal observations and human-verified annotations enable unified evaluation of driving-specific understanding and reasoning supervision for trajectory generation, addressing a gap in existing AD benchmarks. Experiments demonstrate that reasoning supervision improves driving-specific reasoning in VLMs and downstream planning in VLA models, supporting more robust and generalizable AD systems.

\textbf{Limitations.}
The curated four-city distribution does not represent everyday driving or establish global generalization. Human verification cannot eliminate VLM or schema bias or hindsight effects from future-grounded explanations. Counterfactual labels reflect guideline-based judgments rather than observed interventions. Finally, open-loop scoring cannot assess reactive interactions, recovery, or safety under compounding errors, which require closed-loop evaluation.

\clearpage
\bibliographystyle{unsrt}
\bibliography{reference}

\clearpage
\appendix
\input{supp}


\end{document}

%% file: supp.tex
\renewcommand{\thefigure}{S\arabic{figure}}
\renewcommand{\thetable}{S\arabic{table}}
\renewcommand{\theequation}{S\arabic{equation}}
\renewcommand{\theHfigure}{supp.\arabic{figure}}
\renewcommand{\theHtable}{supp.\arabic{table}}
\renewcommand{\theHequation}{supp.\arabic{equation}}

\setcounter{figure}{0}
\setcounter{table}{0}
\setcounter{equation}{0}

\begin{center}
\LARGE \textbf{\textit{nuReasoning} Supplementary Material}
\end{center}

\section{Scenario Evaluation}
\label{eval}

We develop an automatic scenario evaluation pipeline to identify high-value long-tail driving clips from pre-filtered driving logs. The logs are segmented into 30-second clips at 1\,Hz using only front-facing camera views. A VLM-based evaluator is then applied to jointly reason about scene context and ego-vehicle behavior over time.

\begin{figure*}[ht]
    \centering
    \begin{tcolorbox}[
        colback=gray!10,
        colframe=gray!60,
        boxrule=0.4pt,
        arc=2pt,
        left=6pt,
        right=6pt,
        top=6pt,
        bottom=6pt,
        width=\textwidth
    ]
    \footnotesize
    \textbf{Prompt for long-tail scenario evaluation (Part I): Scenario taxonomy and evaluation principle.}

    \vspace{0.4em}
    \textbf{Scenario type definitions:}
    \begin{itemize}
        \item \texttt{normal\_driving}: no unusual agents, objects, infrastructure, or conditions of safety relevance.
        \item \texttt{abnormal\_vehicle\_behavior}: unsafe or unexpected vehicle motion (\textit{e.g.}, cut-ins, swerves, harsh braking, U-turns, wrong-way, reversing into traffic).
        \item \texttt{stationary\_vehicle\_obstruction}: stopped or parked vehicles obstructing or narrowing the lane (\textit{e.g.}, stalled car, double-parking, loading/unloading in lane, doors opening).
        \item \texttt{vulnerable\_road\_user}: pedestrians, cyclists, micro-mobility, wheelchairs, crowds, or children in or near the roadway in risky positions or motions.
        \item \texttt{animal\_on\_road}: animals on or crossing the roadway or immediately adjacent in a way that may affect the ego vehicle.
        \item \texttt{road\_obstacle\_or\_debris}: discrete objects in or near the drivable path (\textit{e.g.}, cargo, tire pieces, branches, cones/barrels acting as obstacles, snow piles intruding into lane).
        \item \texttt{road\_surface\_hazard}: hazardous surface conditions (\textit{e.g.}, potholes, cracks, sinkholes, standing water, flooding, ice, mud, oil).
        \item \texttt{work\_zone\_or\_roadwork}: construction or maintenance altering normal layout (\textit{e.g.}, lane shifts, closures, detours, temporary geometry not in map).
        \item \texttt{traffic\_control\_or\_signage\_anomaly}: unusual or faulty signs, signals, or traffic control (\textit{e.g.}, dark/flashing signals, missing/contradictory signs, manual traffic director, malfunctioning gates).
        \item \texttt{unusual\_traffic\_pattern}: atypical flow patterns (\textit{e.g.}, unexpected gridlock, contraflow, irregular queues/merging, chaotic flows in semi-structured areas).
        \item \texttt{limited\_visibility\_or\_perception\_degradation}: reduced perception due to environment or occlusion (\textit{e.g.}, fog, heavy rain, glare, night, tunnels, strong shadows, occluded intersections, spray on sensors).
        \item \texttt{static\_infrastructure\_constraint}: fixed geometry or clearance limits (\textit{e.g.}, low bridges/tunnels, narrow lanes, overhanging structures/trees constraining vehicle envelope).
        \item \texttt{emergency\_vehicle}: emergency vehicles or scenes requiring special yielding or route changes (active lights/sirens, roadside incident).
        \item \texttt{other}: any other safety-relevant scenario not covered above; briefly describe in the observation field.
    \end{itemize}

    \vspace{0.4em}
    \textbf{Critical evaluation principle: Focus on the ego vehicle's behavior.}
    A scenario is considered high-value long-tail data only if the unusual element(s) cause a clear, non-trivial change in the ego vehicle's behavior, such as braking, stopping, yielding, lane changes, evasive steering, rerouting, waiting, significant speed reduction, or hesitation due to uncertainty or occlusion. The mere presence of unusual or critical objects is not sufficient. If a critical object is visible but does not influence the ego vehicle's actions, the scenario should receive a lower score.
    \end{tcolorbox}
    \caption{Prompt template used for automatic long-tail scenario evaluation and classification. Part I defines the scenario taxonomy and the core principle that clip value is determined primarily by the impact of unusual elements on ego-vehicle behavior.}
    \label{fig:longtail_prompt_1}
\end{figure*}

\textbf{Prompt.}
We design a structured vision-language prompt using Gemini 3.1 Pro to determine whether a clip contains rare or safety-critical driving behavior. The prompt guides the model through chain-of-thought reasoning: (1) describing the scene, (2) summarizing the ego vehicle’s longitudinal and lateral behavior, (3) identifying unusual or rare elements, and (4) assessing whether these elements materially influence the ego vehicle.

A key design principle is the \emph{behavior-centric criterion}: the presence of rare objects alone is insufficient unless it induces a non-trivial behavioral response from the ego vehicle, such as braking, yielding, lane changing, evasive steering, waiting, or significant speed reduction. Scenario categories are defined in \cref{fig:longtail_prompt_1}, and the full prompt template is provided in \cref{fig:longtail_prompt_2}.

The evaluator produces a structured JSON output containing: (1) scene observations, (2) identified unusual elements, (3) safety assessment, (4) rarity assessment, (5) scenario type labels, and (6) a long-tail value score ranging from 1 to 10. Scores of 1-2 correspond to routine driving, 3-4 to noteworthy but behaviorally irrelevant conditions, 5-6 to meaningful behavioral responses, 7-8 to safety-critical interactions, and 9-10 to extremely rare events involving imminent collision risk or highly atypical hazards. Scenario types include normal driving, abnormal vehicle behavior, stationary vehicle obstruction, vulnerable road users, animals, road debris, surface hazards, work zones, traffic-control anomalies, unusual traffic patterns, limited visibility, infrastructure constraints, emergency vehicles, and other safety-relevant cases.

\begin{figure*}[ht]
    \centering
    \begin{tcolorbox}[
        colback=gray!10,
        colframe=gray!60,
        boxrule=0.4pt,
        arc=2pt,
        left=6pt,
        right=6pt,
        top=6pt,
        bottom=6pt,
        width=\textwidth
    ]
    \footnotesize
    \textbf{Prompt for long-tail scenario evaluation (Part II): Reasoning steps, output schema, and scoring rubric.}

    \vspace{0.4em}
    \textbf{Reasoning steps:}
    \begin{enumerate}
        \item \textbf{Scene Description:} Describe the road environment, lighting, weather, and visible agents or objects.
        \item \textbf{Ego Behavior Summary:} Describe the ego vehicle's speed profile, lateral behavior, and right-of-way behavior. Explicitly state if ego behavior is unchanged or typical.
        \item \textbf{Identify Unusual Elements and Ego Influence:} For each unusual element, state whether it changes the ego vehicle's behavior.
        \item \textbf{Assess Safety Criticality:} Evaluate how safety-critical the scenario is for ego decision-making.
        \item \textbf{Determine Rarity:} Assess how rare the scenario is in standard driving datasets.
        \item \textbf{Final Assessment:} Provide a final score driven primarily by ego behavior impact, then by rarity.
    \end{enumerate}

    \textbf{Required JSON output fields:}
    \begin{itemize}
        \item \texttt{observation}: scene description
        \item \texttt{unusual\_elements}: unusual elements and whether they influenced ego
        \item \texttt{safety\_assessment}: safety criticality for ego decision-making
        \item \texttt{rarity\_assessment}: rarity in standard datasets
        \item \texttt{scenario\_types}: one or more scenario types from the predefined list
        \item \texttt{score}: integer from 1 to 10
    \end{itemize}

    \textbf{Scoring rubric:}
    \begin{itemize}
        \item \textbf{1-2}: Normal driving; no unusual elements; ego behavior typical or unchanged.
        \item \textbf{3-4}: Mildly unusual objects or conditions are present, but they do not materially affect ego behavior or safety margin.
        \item \textbf{5-6}: Moderately long-tail scenario; unusual agents, infrastructure, visibility, or traffic patterns require a clear behavioral adjustment from the ego vehicle, such as yielding, slowing, stopping, rerouting within the lane, or maintaining an increased safety buffer, but without immediate near-conflict.
        \item \textbf{7-8}: High-value long-tail scenario; ego must actively manage elevated risk, such as strong braking, evasive lane change, negotiation with vulnerable road users, occluded hazards, or complex multi-agent interaction.
        \item \textbf{9-10}: Extremely rare and safety-critical scenario; imminent collision risk, severe rule conflict, or highly atypical hazards require immediate, non-standard handling.
    \end{itemize}

    \end{tcolorbox}
    \caption{Part II of the automatic long-tail evaluation prompt. It specifies the reasoning procedure, structured JSON output, and score definitions used to assess the value of each clip.}
    \label{fig:longtail_prompt_2}
\end{figure*}

\begin{figure}
    \centering
    \includegraphics[width=\linewidth]{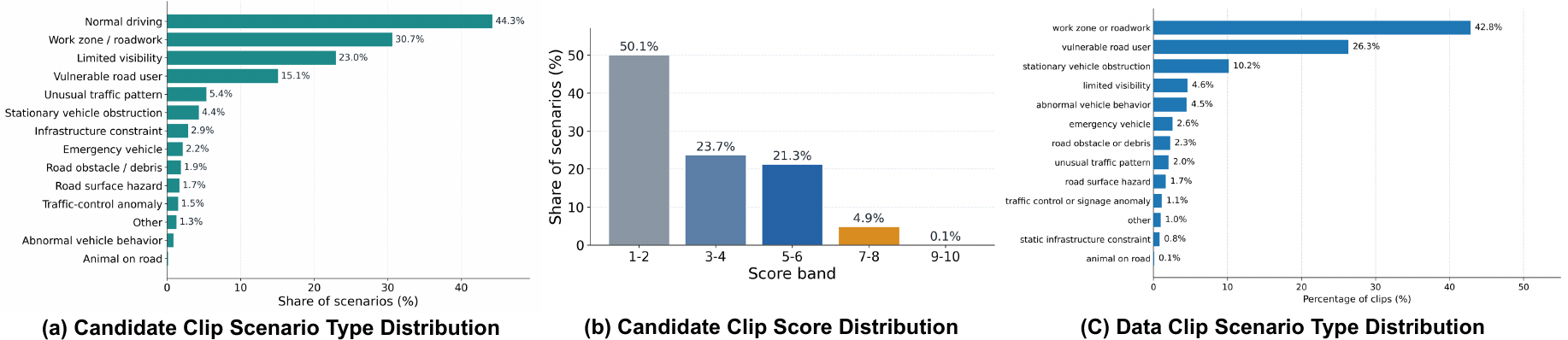}
    \caption{Different stages of scenario selection. (a) Scenario types and (b) Difficulty scores in the initial candidate pool. (c) Scenario types in the final human-verified dataset.}
    \label{fig:scenario_eval_distribution}
    \vspace{-0.3cm}
\end{figure}

\textbf{Results and selection stages.}
We apply the evaluator to more than 100K candidate 30-second clips from engineering-prefiltered logs. Panels (a) and (b) of \cref{fig:scenario_eval_distribution} describe this candidate pool, not the released dataset: normal driving accounts for 44.3\% of candidate scenario types. Only candidates with a difficulty score of 5 or higher are sent for human verification. The 81.72\% confirmation rate is the fraction of these screened-in candidates that experts accept as behaviorally significant long-tail events; rejected candidates are excluded from the final 20K clips. Panel (c) of \cref{fig:scenario_eval_distribution} shows the final scenario-type distribution and has no normal-driving category.

The final score ranges 5-6, 7-8, and 9-10 contain 81.3\%, 18.3\%, and 0.4\% of clips. Under the rubric, the first range requires an unusual agent, road condition, infrastructure element, or visibility issue to cause a clear non-trivial ego response, but does not require an imminent conflict. Higher ranges reflect elevated risk or more atypical hazards. The train, validation, and test splits are stratified by scenario type and difficulty range. These bins describe a deliberately curated benchmark, not the natural frequency of events in routine driving.

\textbf{Human verification and keyframes.}
Experts validate the scenario type, behavioral significance, and decision-critical moment for screened-in candidates, with a second expert double-checking annotations. The keyframe is the point at which ego should decide how to respond: for example, yielding when a pedestrian enters its path, proceeding cautiously under temporary traffic control, or initiating braking or a lane change for a blocked lane. The final clip is extracted from the original continuous log with 10 seconds before and after the selected frame. The screening segment is not a boundary constraint: keyframes near its beginning or end remain eligible when the adjacent log contains sufficient synchronized sensor context.

\textbf{Work-zone prevalence.}
Work zones and roadwork comprise a large share of the curated dataset. This broad category includes lane shifts, narrowed or blocked lanes, workers, maintenance vehicles, and manual traffic controls that materially affect ego behavior. Long-tail curation does not imply a uniform scenario distribution. Appendix~\cref{app:workzone_control} evaluates a reduced-data control that caps the work-zone training fraction at 20\% without changing validation or test data.

\begin{figure*}[t]
    \centering
    \begin{tcolorbox}[
        colback=gray!10,
        colframe=gray!60,
        boxrule=0.4pt,
        arc=2pt,
        left=6pt,
        right=6pt,
        top=6pt,
        bottom=6pt,
        width=\textwidth
    ]
    \footnotesize
    \textbf{Prompt for decision and counterfactual reasoning annotation (Part I): objective, inputs, and critical-component annotation.}

    \vspace{0.4em}
    \textbf{System objective.}
    The model is instructed to act as an autonomous driving safety expert and annotate the realized driving trajectory using camera images, telemetry, and structured spatial context. The output must be a single JSON object with exactly five fields: \texttt{Scene description}, \texttt{Critical components}, \texttt{Driving decision}, \texttt{Reasoning trace}, and \texttt{Counterfactual action analysis}.

    \vspace{0.4em}
    \textbf{Core objective.}
    The prompt requires identifying the \emph{single most restrictive constraint} in the scene according to a strict safety hierarchy, and then explaining how this constraint leads to the realized trajectory. The \textbf{future trajectory is treated as ground truth}; when other inputs appear inconsistent, they are interpreted in a way that justifies the realized path.

    \vspace{0.4em}
    \textbf{Input context.}
    The prompt integrates:
    \begin{itemize}
        \item \textbf{Goal}: high-level routing intent (\texttt{GO\_STRAIGHT}, \texttt{TURN\_LEFT}, \texttt{TURN\_RIGHT});
        \item \textbf{Telemetry}: ego velocity, acceleration, and a 5-second future trajectory in ego coordinates;
        \item \textbf{Spatial context}: object relations, map features, lane structure, and per-camera projections;
        \item \textbf{Camera images}: front, front-left, and front-right temporal sequences.
    \end{itemize}

    \textbf{Goal usage rule.}
    The routing goal provides only high-level intent. When the goal is underspecified, the future trajectory reveals the realized longitudinal and lateral behavior and is used to determine the annotated driving decision.

    \textbf{Conflict-resolution rule.}
    When present, the future trajectory takes precedence over all other modalities. If camera observations, traffic controls, or map context appear inconsistent with the realized path, the scene is interpreted as it must have been for that trajectory to occur.

    \vspace{0.4em}
    \textbf{Scene description.}
    The prompt first requires a concise and objective summary of the driving setting, including road type, traffic condition, traffic control, visibility, and ego placement. This section is descriptive only and must not include reasoning or decision language.

    \vspace{0.4em}
    \textbf{Critical components.}
    The prompt then extracts only those objects, controls, or conditions that \emph{directly force} the realized longitudinal or lateral behavior. Critical components must be identified using only \textbf{past and current} observations, not future camera frames or future trajectory. The schema includes:
    \begin{itemize}
        \item \textbf{Vehicles}: type, relative location, and behavior (\textit{e.g.}, stopped, slowing, cut-in, merge, reversing);
        \item \textbf{Vulnerable road users}: type, relative location, motion state, and gesture if relevant;
        \item \textbf{Traffic controls}: traffic lights and traffic signs with explicit legal effect;
        \item \textbf{Road events and lane features}: curvature, bumps, narrowing, barriers, lane markings, bus lanes, medians;
        \item \textbf{Routing intent and ego state}: route-following constraints and current ego motion state;
        \item \textbf{ODD factors}: weather, visibility, occlusion, construction, emergency vehicles, school-bus rules.
    \end{itemize}

    \textbf{Spatial grounding rule.}
    For vehicles, vulnerable road users, traffic signs, and collision-relevant ODD elements, the prompt requests detailed relative location when the object appears in the structured scene graph; otherwise only a rough distance qualifier is used. Only directly decision-relevant components are retained.

    \vspace{0.4em}
    \textbf{Critical-component example schema.}
    \begin{quote}
    \ttfamily
    \scriptsize
    \{"Critical components": \{"Vehicle 1": \{"Type": "Truck", "Relative location": "18 m in front, right adjacent lane", "Behavior": "Cutting into the ego lane and creating immediate collision risk."\}, "Traffic sign 1": \{"Type": "Stop", "Relative location": "Front right, directly ahead", "Impact": "Complete stop required before proceeding."\}\}\}
    \end{quote}
    \end{tcolorbox}
    \caption{Prompt template for decision and counterfactual reasoning annotation. Part I defines the task objective, multimodal inputs, conflict-resolution rule, and the schema used to identify critical scene components that directly constrain the realized trajectory.}
    \label{fig:decision_prompt_1}
    \vspace{-0.4cm}
\end{figure*}

\section{Reasoning Annotation}
\label{reasoning}
\label{app:annotation}

We augment each target frame with structured reasoning annotations that explain the realized ego trajectory in terms of scene constraints, driving actions, and counterfactual alternatives. The annotation pipeline integrates synchronized multi-view camera observations, ego telemetry, map context, object-level spatial relations, and the future ego trajectory.

\subsection{Spatial Reasoning Annotation}
Gemini 3 Flash identifies driving-relevant objects in synchronized multi-view images. The 3D categories, track identities, boxes, positions, orientations, and velocities are produced separately by an internal offline perception and tracking pipeline using aggregated multi-frame point clouds. Camera calibration projects each 3D box into the corresponding view; overlap-based association links reliable 2D detections to 3D tracks. Only matched detections receive metric positions, distances, and velocities. Unmatched detections remain image-level observations, while ambiguous or conflicting matches are corrected or left unassociated. Consequently, metric spatial attributes are computed from sensor-derived tracks and map geometry rather than generated as free-form VLM values.

We further enrich each matched object with temporal and interaction attributes. Future object trajectories are obtained from subsequent annotations when available, or approximated using a constant-velocity model otherwise. These trajectories support path-intersection, time-to-collision, footprint-overlap, minimum-distance, and conflict-type labels. Nearby lane centerlines and crosswalks are represented in the ego frame and projected into camera views. Future-derived attributes are annotation targets, not future observations supplied to evaluated models; the constant-velocity fallback can limit the reliability of motion labels in complex interactions.

\subsection{Driving Reasoning Annotation}

The driving reasoning stage converts spatial context into a structured explanation of the ego vehicle’s behavior. The annotator takes as input ego velocity and acceleration, high-level navigation commands, temporal camera context, structured spatial annotations, and the future ego trajectory. The future trajectory serves as the reference for action selection, while past and current observations are used to identify the constraints that explain this behavior.

Each driving annotation consists of four components: \emph{scene description}, \emph{critical components}, \emph{driving decision}, and \emph{reasoning trace}. Critical components include only the objects or conditions that directly constrain the maneuver, such as traffic lights, stop signs, pedestrians, leading vehicles, lane blockages, road geometry, construction, visibility limitations, or route intent.

The driving decision is represented by one longitudinal action and one lateral action selected from a fixed taxonomy. Action selection must match the ground-truth future trajectory, while a priority hierarchy is used solely to explain the causal factors behind the decision: (1) collision avoidance, (2) regulatory compliance, (3) geometry and comfort, and (4) efficiency. The prompt specification is shown in \cref{fig:decision_prompt_1,fig:decision_prompt_2}.

The reasoning trace is expressed as a concise, single-sentence causal chain that identifies the dominant constraint, explains how it limits feasible motion, and justifies the selected action.

\begin{figure*}[ht]
    \centering
    \begin{tcolorbox}[
        colback=gray!10,
        colframe=gray!60,
        boxrule=0.4pt,
        arc=2pt,
        left=6pt,
        right=6pt,
        top=6pt,
        bottom=6pt,
        width=\textwidth
    ]
    \footnotesize
    \textbf{Prompt for decision and counterfactual reasoning annotation (Part II): driving decision, reasoning trace, and counterfactual analysis.}

    \vspace{0.4em}
    \textbf{Driving decision.}
    The prompt requires selecting exactly one longitudinal action and one lateral action to match the realized future trajectory.

    \textbf{Longitudinal action set.}
    The nine longitudinal actions are:
    \emph{Remain stopped},
    \emph{Quickly come to a stop},
    \emph{Gently come to a stop},
    \emph{Slow down quickly},
    \emph{Slow down gently},
    \emph{Quickly accelerate (speed up)},
    \emph{Gently accelerate (speed up)},
    \emph{Maintain speed}, and
    \emph{Reverse}.

    \textbf{Lateral action set.}
    The seven lateral actions are:
    \emph{Slightly move left in the lane},
    \emph{Slightly move right in the lane},
    \emph{Left lane change},
    \emph{Right lane change},
    \emph{Turn left},
    \emph{Turn right}, and
    \emph{No lateral action}.

    \textbf{Action-selection rule.}
    The selected actions must first match the ground-truth future motion. The priority hierarchy is then used only to explain \emph{why} that trajectory was taken. A special constraint is that if the longitudinal action is \emph{Remain stopped}, the lateral action must be \emph{No lateral action}.

    \vspace{0.4em}
    \textbf{Priority hierarchy.}
    The prompt organizes causal justification using a four-level hierarchy:
    \begin{enumerate}
        \item \textbf{Collision avoidance}: blocked paths, crossing agents, cut-ins, and immediate hazards;
        \item \textbf{Regulatory compliance}: red lights, stop signs, yield conditions, school-bus and rail-crossing rules;
        \item \textbf{Geometry and comfort}: curves, bumps, narrow gaps, poor road surface, and limited visibility;
        \item \textbf{Efficiency}: normal unconstrained progress.
    \end{enumerate}

    \vspace{0.4em}
    \textbf{Reasoning trace.}
    The prompt then produces a short, single-sentence logic chain linking the dominant constraint to the realized maneuver. It identifies the highest-priority factor, explains why it constrains motion, and states the chosen action.

    \vspace{0.4em}
    \textbf{Counterfactual action analysis.}
    Finally, the prompt evaluates plausible alternative longitudinal-lateral action pairs using only \textbf{past and current} scene information. It does \emph{not} enumerate all combinations, but considers only meaningful alternatives allowed by the current road layout, constraints, and ego state. The alternatives are divided into:
    \begin{itemize}
        \item \textbf{Alternative actions}: plausible and safe action pairs only;
        \item \textbf{Top safety-critical actions}: the most dangerous or clearly suboptimal alternatives.
    \end{itemize}

    Each candidate pair is annotated with a risk level and a short causal explanation. The allowed risk labels are \texttt{Safe} for alternative actions, and \texttt{Suboptimal} or \texttt{Unsafe} for top safety-critical actions.
    
    \vspace{0.4em}
    \textbf{Final output format.}
    The final response must contain exactly five fields:
    \begin{itemize}
        \item \texttt{Scene description}
        \item \texttt{Critical components}
        \item \texttt{Driving decision}
        \item \texttt{Reasoning trace}
        \item \texttt{Counterfactual action analysis}
    \end{itemize}
    and must be returned as a single valid JSON object with no surrounding text or markdown.
    \end{tcolorbox}
    \caption{Part II of the decision-reasoning prompt. It defines the exact longitudinal and lateral action sets, the priority hierarchy used for causal justification, concise reasoning-trace generation, and structured counterfactual action analysis.}
    \label{fig:decision_prompt_2}
    \vspace{-0.4cm}
\end{figure*}

\paragraph{Interpretation of future-grounded labels.}
The future-trajectory precedence rule grounds labels in the realized behavior, not a uniquely valid decision or the driver's internal reasoning. Critical components are requested from past/current observations, but future conditioning can still bias an explanation. The displayed prompts document pre-annotation: human review corrects unsupported labels and removes unresolved ambiguous cases. A benchmark score should therefore be interpreted as agreement with a verified explanation of demonstrated behavior, not evidence that the model has recovered a causal decision process.

\paragraph{Visual cues and residual style bias.}
Temporal multi-view images can provide brake lights, turn signals, hazard lights, pedestrian gestures, and traffic-controller signals in addition to metric motion. These cues may inform decision-critical components, but are not currently stored as separate structured attributes. Both scene selection and reasoning annotation use VLM pre-annotations with different prompts, schemas, and acceptance criteria, followed by human validation. Shared pre-annotator and schema biases can nevertheless remain. No independently human-written trace subset is available to isolate style effects; Text F1 and ROUGE-L remain auxiliary rather than causal-correctness measures.

\subsection{Counterfactual Reasoning Annotation}

The counterfactual annotation stage evaluates plausible alternative action pairs under the same scene conditions. Rather than exhaustively enumerating all longitudinal–lateral combinations, it focuses on meaningful alternatives that are feasible given the current road layout, constraints, and ego state. These alternatives are categorized into: (1) safe and plausible actions, and (2) safety-critical actions that are suboptimal or unsafe. Each candidate is assigned a risk label (\texttt{Safe}, \texttt{Suboptimal}, or \texttt{Unsafe}) along with a brief causal explanation. Safety-critical alternatives are further ranked by their potential risk, such as collision likelihood, rule violation, or degradation of driving safety. This process provides explicit supervision not only for the chosen maneuver but also for why other actions are less appropriate.

\paragraph{Risk-label criteria.}
The shared priority hierarchy is collision avoidance, regulatory compliance, road geometry and comfort, then efficiency. \texttt{Safe} means a feasible alternative without an additional safety or rule-compliance concern. \texttt{Suboptimal} means a feasible alternative that unnecessarily reduces safety margin, comfort, efficiency, or route progress. \texttt{Unsafe} means a concrete collision, rule-violation, blocked-path, or drivable-area risk. For example, the annotations in \cref{fig:annotation} label maintaining speed behind a decelerating lead vehicle as \texttt{Suboptimal}, and changing right into a stationary truck and construction barrels as \texttt{Unsafe}. Each alternative is judged using the same scene context and audited by experts; unresolved boundary cases are excluded. These labels are guideline-based judgments of unexecuted actions, not measured outcomes of interventions, and some boundaries remain driving-policy dependent.

\subsection{Human Verification}
Annotators are trained in driving-scene understanding, behavior interpretation, and safety assessment. Guideline-based calibration covers the action taxonomy, critical-agent identification, and risk assessment. A first expert reviews decision and counterfactual pre-annotations, and a second independently audits the reviewed labels. Errors in actions, critical components, rationales, alternative actions, and risk levels are corrected; cases that remain ambiguous are removed rather than assigned a forced label.

\cref{tab:annotation_quality} separates the quality of automatic pre-annotations from the final labels. Scenario confirmation is measured among screened-in candidates; decision and counterfactual agreement is measured before correction. These quantities are not final-label accuracy or independent inter-annotator agreement. Spatial annotations use geometric association and sensor-derived metric attributes, so a free-form reasoning agreement rate is not directly comparable. Human verification reduces, but cannot eliminate, systematic annotation-model and schema biases.

\subsection{QA Generation}

We transform the structured annotations into VQA examples grounded in the \texttt{Spatial}, \texttt{Driving}, and \texttt{Counterfactual} components. The QA generation module produces both multiple-choice and numerical questions. Spatial questions cover object identity, ego-relative 3D position, distance, velocity, 2D bounding-box coordinates, image localization, multi-view correspondence, lane and crosswalk geometry, future motion, and interaction outcomes. Driving questions focus on the joint longitudinal–lateral decision, reasoning trace, traffic rules, critical scene elements, and overall scene understanding. Counterfactual questions assess the safety and risk level of alternative actions.

Each QA instance is augmented with metadata specifying its temporal scope (current frame, historical context, or future prediction) and view requirements. Optional language-model-based rephrasing is applied to diversify question phrasing and generate plausible distractors while preserving the correctness of the ground-truth answers.

\paragraph{Training-pair sampling.}
The complete annotations yield more than 10M QA pairs; the 167K-pair training subset is sampled hierarchically by capability group, fine-grained subcategory, answer format, scenario type, and difficulty range. Per-clip caps and removal of near-duplicate templates prevent frequent clips and question forms from dominating the selected subset. Question-level task metadata determines which past/current observations are available; future annotation targets are not inserted into the evaluated model's input.

\begin{figure}[ht]
    \centering
    \includegraphics[width=\linewidth]{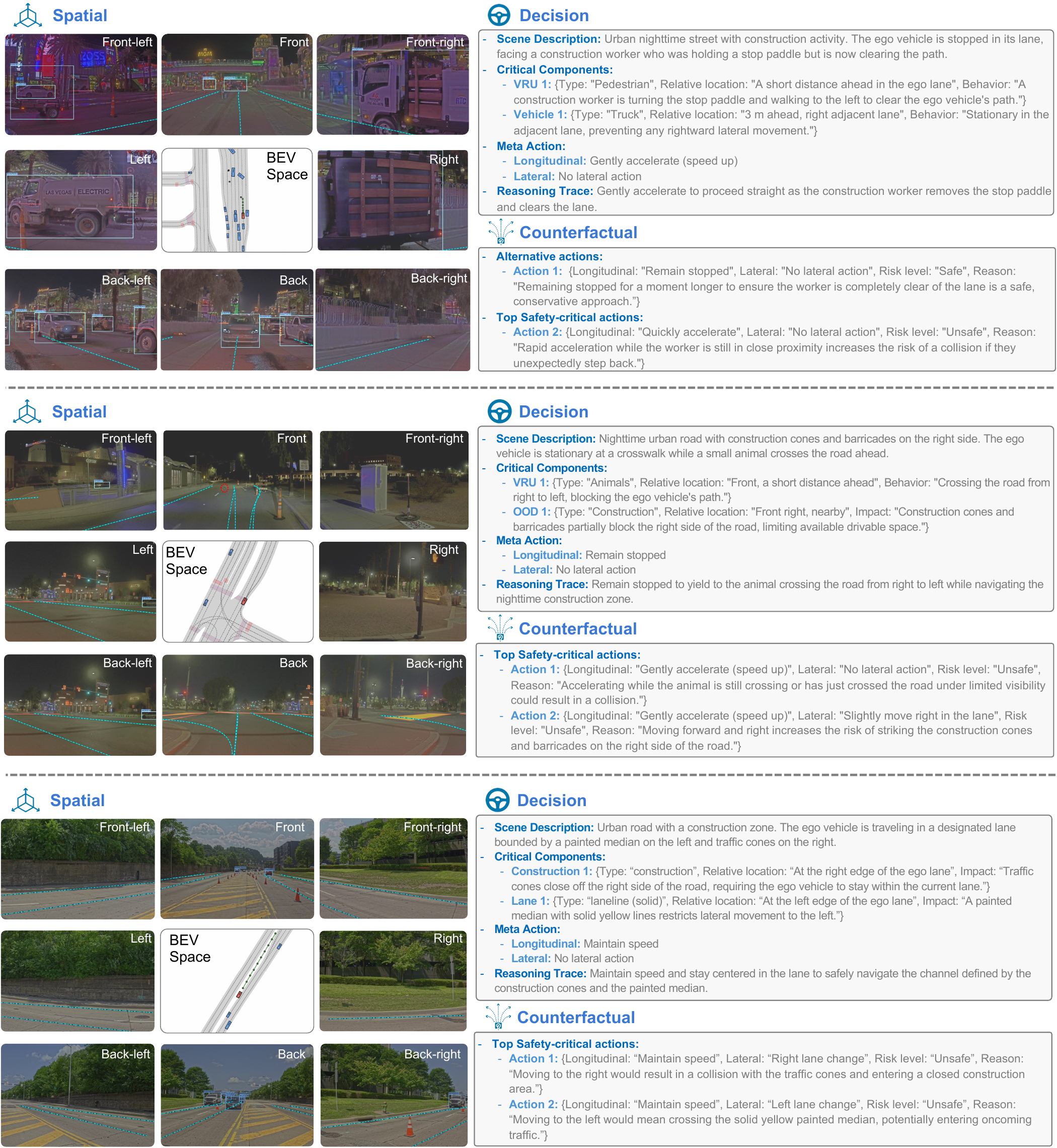}
    \caption{Additional examples of nuReasoning annotations across diverse driving scenarios.}
    \label{fig:additional_annotations}
    \vspace{-0.4cm}
\end{figure}

Additional annotation examples are provided in \cref{fig:additional_annotations}.

\section{nuVLA Model Details}
\label{app:vla}

\subsection{Model}

\textbf{nuVLA architecture.}
nuVLA is a vision-language-action model for trajectory-level planning in autonomous driving. It contains two trainable components: a Qwen3-VL-2B-Instruct vision-language backbone and a flow-matching Diffusion Transformer (DiT) action expert. The VLM encodes multi-view, multi-frame camera observations together with a natural-language driving prompt, and is supervised to produce structured driving reasoning. The action expert consumes the VLM prompt-side hidden states, ego dynamics, and ego motion history to generate a future ego trajectory for open-loop planning evaluation.

\textbf{Multi-view image input.}
For each scene, we provide synchronized camera images from eight surrounding views: front, front-left, front-right, left, right, back, back-left, and back-right. The number of temporal input steps is configurable. In our default setting, each sample uses one history timestep together with the current timestep, resulting in up to $2 \times 8=16$ images per scene. Users may adjust the temporal window according to the model design, compute budget, and task requirements. All input images are resized to $448 \times 448$ before being passed to the Qwen processor. Each image is inserted into the chat prompt with an explicit textual tag indicating both its relative time and camera view, e.g., ``$t=-1$, front camera'' or ``$t=0$ (current), front-left camera''. This design lets the VLM associate visual evidence with both spatial viewpoint and temporal context, rather than treating the image list as an unordered collection.

The prompt describes the model role as an autonomous driving assistant, states that the inputs are multi-view multi-frame observations, and includes the mission command, such as \textit{LANE\_FOLLOW}. The prompt then requests structured reasoning in the configured format. Our default format is \textit{spatial-driving}: the model predicts both spatial scene understanding and driving-oriented reasoning.

\textbf{Reasoning supervision.}
The spatial reasoning target includes a scene/layout summary and object-centric spatial relations. For nearby or relevant agents, the target may include object category, ego-frame 3D center, per-camera 2D bounding boxes, distance to the ego vehicle, predicted future conflict indicators such as time-to-collision and minimum distance, and short-horizon future object trajectories. Objects are ordered by distance to the ego vehicle and truncated to a fixed maximum number of items.
The driving reasoning target contains a driving-oriented scene description, critical components, a structured driving decision, and a reasoning trace. The decision is decomposed into longitudinal and lateral components, such as maintaining speed, decelerating, stopping, lane following, or steering behavior. In formats that include counterfactual supervision, the target additionally contains alternative safe actions and top safety-critical or unsafe actions, together with risk rationales. This reasoning supervision encourages the VLM to expose planning-relevant intermediate factors rather than only producing a final action.

\textbf{Vision-language backbone.}
The VLM backbone is initialized from Qwen3-VL-2B-Instruct and fine-tuned using LoRA adapters. We apply LoRA to the attention and MLP projection modules, including query, key, value, output, gate, up, and down projections. The default LoRA rank is $32$, the scaling factor is $64$, and dropout is $0.1$. The model uses bfloat16 training and FlashAttention when available. We use the last-layer hidden states of the VLM as vision-language conditioning features. During training, only the prompt-side hidden states, i.e., the user turn containing image tokens, camera/time labels, mission command, and reasoning instruction, are passed to the action expert.

\textbf{Action expert.}
The action expert follows a flow-matching DiT design. It predicts an ego-frame future trajectory
$
\hat{\tau} =
\{(\hat{x}_t,\hat{y}_t,\hat{\theta}_t)\}_{t=1}^{T},
$
where $T=10$, corresponding to a 5-second horizon at 2 Hz. The trajectory is represented as cumulative future waypoints in the current ego frame, with each waypoint containing longitudinal displacement, lateral displacement, and heading. Before flow-matching training, trajectory dimensions are normalized by fixed scales $(50.0, 20.0, \pi)$ for longitudinal position, lateral position, and heading.

The action expert receives three conditioning sources: VLM hidden states, current ego dynamics, and ego motion history. The ego dynamics are represented as $(v_x,v_y,a_x,a_y)$. The history trajectory contains past ego poses in the ego frame and is flattened together with ego dynamics into a state token. Future action tokens are produced by an action encoder that combines the noisy trajectory with a sinusoidal timestep embedding.

\textbf{Diffusion Transformer.}
The DiT operates on the concatenation of one state token and $T$ action tokens. It uses hidden dimension $512$, $12$ transformer layers, $8$ attention heads, and dropout $0.1$. Transformer blocks alternate between cross-attention to VLM features and self-attention over state/action tokens. Each block is conditioned on the flow-matching timestep through adaptive layer normalization. The action-token outputs are decoded by an MLP into trajectory velocity predictions.

\subsection{Training}

nuVLA is trained with a language reasoning objective and a flow-matching action objective. Given multi-view observations $I$, driving prompt $q$, and target reasoning text $y$, the VLM is supervised by
\begin{equation}
\mathcal{L}_{\mathrm{reason}}
=
-\sum_i \log p_\phi(y_i \mid y_{<i}, I, q),
\end{equation}
where the loss is computed only on assistant-response tokens.

For the action objective, let $x_1$ be the normalized ground-truth future trajectory and let $\epsilon \sim \mathcal{N}(0,I)$ be Gaussian noise. We sample a flow-matching time $t$ from a Beta distribution and form
\begin{equation}
x_t = (1-t)\epsilon + t x_1.
\end{equation}
The target velocity field is
\begin{equation}
v^\star = x_1 - \epsilon.
\end{equation}
The action expert predicts $v_\theta(x_t,t,c)$, where $c$ contains the VLM prompt features, ego dynamics, and ego history trajectory. It is trained with
\begin{equation}
\mathcal{L}_{\mathrm{action}}
=
\left\|
v_\theta(x_t,t,c) - v^\star
\right\|_2^2.
\end{equation}
The full objective is
\begin{equation}
\mathcal{L}
=
\mathcal{L}_{\mathrm{reason}}
+
\lambda_{\mathrm{action}}\mathcal{L}_{\mathrm{action}},
\end{equation}
with $\lambda_{\mathrm{action}}=1.0$.

We train for 10 epochs with batch size 2 per GPU, gradient accumulation of 4, VLM learning rate $5\times10^{-5}$, action-expert learning rate $1\times10^{-4}$, weight decay $0.01$, warmup ratio $0.05$, and gradient clipping at norm $1.0$. The planning horizon is 5 seconds at 2 Hz, giving 10 future waypoints.

\subsection{Inference}

At inference time, nuVLA encodes the multi-view camera observations and driving prompt with the Qwen3-VL backbone. The resulting prompt-side hidden states, together with ego velocity, acceleration, and history trajectory, condition the action expert. The action expert initializes the future trajectory from Gaussian noise, $x_0 \sim \mathcal{N}(0,I)$, and performs Euler integration of the velocity field:
\begin{equation}
x_{k+1}=x_k + \Delta t\,v_\theta(x_k,t_k,c),
\end{equation}
where $\Delta t=1/K$. We use $K=5$ inference steps for planning evaluation. The final sample is denormalized using $(50.0,20.0,\pi)$, and heading angles are wrapped to $[-\pi,\pi]$. The resulting 5-second ego-frame trajectory is passed to the planning benchmark for collision, drivable-area, progress, comfort, human-likeness, and overall planning-score evaluation.

\subsection{Controlled Training and Baseline Comparability}
\label{app:baseline_protocol}
All trainable in-domain baselines are trained on the nuReasoning training split and evaluated on the same test set and 5-second planning horizon. UniAD uses planning and BEV supervision, DiffusionDrive uses planning supervision, and AutoVLA uses planning and decision supervision. Alpamayo-1.5 is evaluated zero-shot and is not an in-domain retrained control. Their architectural representations, input configurations, and training budgets are not fully harmonized, so \cref{vla_results} is a system-level comparison rather than an architecture-only ablation.

The controlled nuVLA variants use the same backbone, action expert, temporal multi-view inputs, planning targets, split, initialization protocol, optimizer settings, and inference procedure. Only auxiliary supervision differs. All planning-only samples come from the same annotated-clip training pool; removing reasoning supervision does not introduce extra planning data. Reasoning targets supervise the assistant response, while the standard action expert uses only prompt-side hidden states rather than ground-truth reasoning-response states. These controls make the within-nuVLA comparisons the primary evidence for the utility of reasoning supervision.

\section{Planning Benchmark}
\label{app:planning_metrics}

\paragraph{Planning metric details.}
For each evaluated clip, we select a key frame and evaluate the predicted future ego trajectory
$\hat{\tau}=\{(\hat{x}_t,\hat{y}_t,\hat{\psi}_t)\}_{t=0}^{T}$ over a 5-second horizon at $\Delta t=0.1$ seconds, yielding 51 poses including the current pose. Trajectories are represented in the global coordinate frame using ego position and heading. Predictions are interpolated onto this benchmark grid and compared against the ground-truth future ego trajectory, the intended route, the static map, and future annotations of surrounding objects.

\paragraph{Non-at-fault collision.}
The collision score $s_{\text{NC}}$ measures whether the predicted ego trajectory causes an at-fault contact with surrounding objects. At each predicted waypoint, we construct the oriented ego bounding box using the ego footprint and rear-axle-to-center offset, and test the bird's-eye-view polygon overlap with annotated objects at the corresponding timestep. Any positive overlap is treated as a contact. Objects categorized as vehicles, pedestrians, bicycles, or related dynamic classes are treated as agents; cones, barriers, and other non-agent objects are treated as static objects, while ambiguous ``other'' classes are ignored.

Each contact is classified using relative geometry and motion. Contacts while the ego is stopped are non-at-fault. Contacts with stopped tracks and active front contacts are at fault. Active rear contacts are non-at-fault. Active lateral contacts are at-fault only when the ego is outside the drivable area or spans multiple lanes. Once a track has been processed as a contact, subsequent overlaps with the same track are ignored. If an at-fault collision occurs with a dynamic agent, we set $s_{\text{NC}}=0$. If the at-fault collision is with a static object, we assign the partial score $s_{\text{NC}}=0.5$. If no at-fault collision occurs, $s_{\text{NC}}=1$.

\paragraph{Drivable-area compliance.}
The drivable-area score $s_{\text{DA}}$ evaluates whether the predicted ego footprint remains within the valid road surface. We build the drivable region $\mathcal{D}$ as the union of lane, intersection, and road-block polygons from the static map, expanded by a $0.5$ m buffer. For each predicted waypoint, all four corners of the ego bounding box are checked against $\mathcal{D}$. Timesteps whose ego footprint lies completely outside the known map extent are skipped, so missing map coverage is not penalized. The score is binary:
\begin{equation}
s_{\text{DA}} =
\mathbb{I}\left[
\forall t \in \mathcal{T}_{\mathrm{map}},\
\mathrm{corners}(\hat{x}_t,\hat{y}_t,\hat{\psi}_t)
\subseteq \mathcal{D}
\right],
\end{equation}
where $\mathcal{T}_{\mathrm{map}}$ denotes timesteps evaluated within map coverage. 

\paragraph{Ego progress.}
Ego progress $s_{\text{EP}}$ measures whether the ego vehicle advances along the intended route. We construct a route polyline from the mission goal route path, interpolate it, and project the starting point and final predicted point onto the route. The predicted progress is
$p_{\mathrm{pred}}=\max(0, q_{\mathrm{pred}}-q_0)$, where $q_0$ and $q_{\mathrm{pred}}$ are the route projection distances of the initial and final predicted points. The ground-truth progress $p_{\mathrm{gt}}$ is computed analogously from the final ground-truth point. The score is
\begin{equation}
s_{\text{EP}}
=
\begin{cases}
1, & p_{\mathrm{gt}} < 0.1,\\
\mathrm{clip}\left(\frac{p_{\mathrm{pred}}}{p_{\mathrm{gt}}}, 0, 1\right), & \text{otherwise}.
\end{cases}
\end{equation}

\paragraph{Comfort.}
The comfort score $s_{\text{Comf}}$ measures whether the predicted trajectory satisfies fixed smoothness constraints. The trajectory is subsampled to a $0.5$ second cadence, while preserving the final point, and differentiated with Savitzky-Golay filters. We estimate longitudinal acceleration, lateral acceleration, acceleration-magnitude jerk, longitudinal jerk, yaw rate, and yaw acceleration. The score is binary and equals $1$ only if all quantities remain within their configured bounds over the horizon. Otherwise, $s_{\text{Comf}}=0$.

\paragraph{Human-likeness.}
Human-likeness $s_{\text{HL}}$ measures similarity between the predicted and demonstrated future trajectories using final displacement error. Over the shared horizon, we compute per-step displacement errors and report both ADE and FDE:
\begin{equation}
\mathrm{ADE}
=
\frac{1}{T}
\sum_{t=1}^{T}
\left\|
(\hat{x}_t,\hat{y}_t) - (x_t,y_t)
\right\|_2,
\quad
\mathrm{FDE}
=
\left\|
(\hat{x}_T,\hat{y}_T) - (x_T,y_T)
\right\|_2.
\end{equation}
The human-likeness score is derived from FDE with a smoothstep penalty. Let $a=1.0$ m, $b=8.0$ m, and
$u=(\mathrm{FDE}-a)/(b-a)$. Then
\begin{equation}
s_{\text{HL}} =
\begin{cases}
1, & \mathrm{FDE} \le a,\\
0, & \mathrm{FDE} \ge b,\\
1 - u^2(3-2u), & \text{otherwise}.
\end{cases}
\end{equation}
The reason to use this instead of a linear penalty is that it avoids harsh changes near the thresholds. Small errors just above 1 meter are not punished too abruptly, and scores taper gently as they approach the failure threshold. 

\paragraph{Aggregation.}
The overall planning score combines the above metrics as
\begin{equation}
\mathrm{NPS}
=
s_{\text{NC}} s_{\text{DA}}
\frac{
w_{\text{EP}}s_{\text{EP}}
+
w_{\text{Comf}}s_{\text{Comf}}
+
w_{\text{HL}}s_{\text{HL}}
}{
w_{\text{EP}}+w_{\text{Comf}}+w_{\text{HL}}
}.
\end{equation}

The score is computed per clip before averaging, rather than multiplying dataset-mean sub-scores. Dynamic-agent at-fault collisions and drivable-area violations yield zero through their gates, while an at-fault static-object collision receives the documented partial NC score of 0.5. We report mean NPS and each normalized sub-score on a 0-100 scale, together with mean ADE/FDE in meters. Under the default weights $(0.3,0.2,0.5)$, the denominator is one, giving \cref{eq:nps}.

\subsection{Weight Sensitivity and Interpretation}
\label{app:metric_sensitivity}
The default weights are a domain-motivated operating point, not a uniquely optimal choice: human-likeness emphasizes the demonstrated context-dependent response, progress rewards effective route completion, and comfort rewards smoothness without dominating decision quality. NC and DA remain multiplicative rather than additive safety terms. \cref{tab:weight_sensitivity} reports the three-seed weight analysis alongside the same supervision configurations as \cref{tab:planning_three_seeds}. Every reasoning-supervised model exceeds planning only, and S+D+CF ranks first under all five tested choices. The ordering of close intermediate variants is not invariant: under progress-focused weights, S+D slightly exceeds D.

\begin{table}[t]
\centering
\caption{NPS weight sensitivity (mean $\pm$ standard deviation over three seeds). Column tuples are $(w_{\mathrm{EP}},w_{\mathrm{Comf}},w_{\mathrm{HL}})$; NC and DA are unchanged. The default column matches the three-seed supervision ablation.}
\label{tab:weight_sensitivity}
\footnotesize
\setlength{\tabcolsep}{4.3pt}
\begin{tabular}{lccccc}
\toprule
Reasoning & Default & Balanced & Progress & Comfort & HL \\
& 0.30/0.20/0.50 & 0.33/0.33/0.34 & 0.50/0.20/0.30 & 0.25/0.50/0.25 & 0.20/0.20/0.60 \\
\midrule
None & $64.52\pm0.31$ & $70.15\pm0.35$ & $69.68\pm0.36$ & $74.43\pm0.37$ & $61.94\pm0.32$ \\
D & $71.51\pm0.41$ & $76.68\pm0.36$ & $76.41\pm0.30$ & $80.53\pm0.36$ & $69.06\pm0.48$ \\
D+CF & $72.33\pm0.06$ & $77.52\pm0.06$ & $77.33\pm0.10$ & $81.33\pm0.09$ & $69.83\pm0.11$ \\
S & $70.18\pm0.21$ & $75.07\pm0.14$ & $74.98\pm0.05$ & $78.60\pm0.15$ & $67.78\pm0.28$ \\
S+D & $71.19\pm0.34$ & $76.58\pm0.33$ & $76.43\pm0.37$ & $80.50\pm0.32$ & $68.57\pm0.38$ \\
S+D+CF & $\mathbf{72.86\pm0.07}$ & $\mathbf{78.04\pm0.04}$ & $\mathbf{77.87\pm0.10}$ & $\mathbf{81.82\pm0.06}$ & $\mathbf{70.36\pm0.13}$ \\
\bottomrule
\end{tabular}
\end{table}

\paragraph{Scope of the score.}
NPS summarizes log-conditioned planning quality, not safety under reactive interactions. HL is derived from the recorded endpoint and is therefore not an independent human-preference judgment. Agreement between NPS and HL is also not independent validation because HL appears inside NPS. Reporting ADE and the individual sub-scores exposes trade-offs, but a single demonstrated trajectory does not enumerate all safe alternatives. Repeated planning on logged observations would remain non-reactive; closed-loop validation requires reconstructed scenes and agents that respond to ego's actions.

\section{Reasoning Benchmark}
\label{app:eval_protocol}

This section provides the evaluation details for the VLM reasoning benchmark introduced in \cref{sec:reasoning_evaluation}. The benchmark is organized around four main categories: geometry, motion, driving, and counterfactual reasoning. This grouping is used only for analysis and reporting; each individual question is still evaluated with the metric appropriate for its answer format. The question categories and counts for evaluation are shown in \cref{QA_test}.

\begin{table*}[t]
\centering
\small
\renewcommand{\arraystretch}{1.08}
\setlength{\tabcolsep}{5pt}
\caption{Question categories in four main reasoning capabilities. Counts are from the private testing set manifest (16{,}437 evaluable samples in total).}
\label{tab:app_reasoning_categories}
\resizebox{\textwidth}{!}{
\begin{tabular}{l|c|p{0.72\textwidth}}
\toprule
Category & Count & Subcategories \\
\midrule
Geometry & 6,578 & \texttt{object\_xy\_position}, \texttt{object\_relative\_position}, \texttt{object\_distance}, \texttt{distance\_from\_2d\_bbox}, \texttt{bbox\_center\_coordinates\_1000}, \texttt{normalized\_bbox\_coordinates\_1000}, \texttt{ego\_position\_from\_2d\_bbox}, \texttt{image\_region\_from\_3d\_object}, \texttt{camera\_view\_for\_3d\_object}, \texttt{closest\_object\_type}, \texttt{multiview\_camera\_pair}, \texttt{current\_lane\_coordinates\_2d\_1000}, \texttt{nearby\_lane\_count}, \texttt{nearby\_crosswalk\_count}, \texttt{nearby\_crosswalk\_coordinates\_2d\_1000} \\
Motion & 3,172 & \texttt{future\_motion\_label}, \texttt{future\_xy\_trajectory}, \texttt{future\_path\_intersection}, \texttt{future\_time\_to\_conflict\_s}, \texttt{conflict\_prediction}, \texttt{motion\_relation}, \texttt{object\_speed} \\
Driving & 2,619 & \texttt{driving\_decision}, \texttt{driving\_reasoning\_trace}, \texttt{traffic\_regulation\_light\_state}, \texttt{traffic\_regulation\_signal} \\
Counterfactual & 4,068 & \texttt{unsafe\_action\_identification}, \texttt{action\_risk\_assessment} \\
\bottomrule
\end{tabular}
}
\label{QA_test}
\end{table*}

\textbf{Label reliability.}
nuReasoning reduces reliance on purely free-form text evaluation by using structured answer formats whenever possible. Most questions are evaluated against explicit labels such as object positions, distances, object counts, map-derived lane and crosswalk annotations, traffic-light states, future motion categories, and trajectory or coordinate targets. Counterfactual and driving-decision questions are evaluated as multiple-choice problems with fixed answer sets, which avoids open-ended language judgment. Free-form reasoning traces appear only in the \texttt{driving\_reasoning\_trace} subcategory and are reported with reference-based text metrics as auxiliary signals rather than as the primary basis for our conclusions. The test split is held out from fine-tuning. Overall, this design makes the evaluation less sensitive to language style matching and emphasizes structured, metric-grounded driving reasoning.

\textbf{Output parsing.}
Model outputs are parsed deterministically according to the expected answer format. For multiple-choice questions, we normalize the response to either the option label or the option text. For scalar numerical questions, we extract a numeric value and compare it with the annotated target. For coordinate and trajectory questions, we parse structured numeric outputs into points or waypoint sequences before evaluation. If an output cannot be parsed into the required format, it is counted as incorrect for accuracy, tolerance-based, and hit-rate metrics; error-only summaries exclude predictions only when no valid numeric prediction exists.

\textbf{Choice and numerical metrics.}
For multiple-choice questions, we report choice accuracy
\begin{equation}
\mathrm{ACC}_{\mathrm{choice}}
=
\frac{1}{N_{\mathrm{c}}}
\sum_{i=1}^{N_{\mathrm{c}}}
\mathbf{1}\!\left[\hat{y}_i = y_i\right],
\end{equation}
where $\hat{y}_i$ and $y_i$ denote the predicted and ground-truth choices. For numerical questions, we report tolerance-based numerical accuracy
\begin{equation}
\mathrm{ACC}_{\mathrm{num}}
=
\frac{1}{N_{\mathrm{n}}}
\sum_{i=1}^{N_{\mathrm{n}}}
\mathbf{1}\!\left[
|\hat{v}_i - v_i| \leq \epsilon_i
\right],
\end{equation}
where $\hat{v}_i$ is the predicted scalar value, $v_i$ is the ground-truth value, and $\epsilon_i$ is a task-specific tolerance. 

\textbf{Coordinate and trajectory metrics.}
Coordinate metrics are computed in the coordinate system required by each question, including normalized image coordinates, bounding-box coordinates, and ego-frame object coordinates. Given a predicted coordinate $\hat{\mathbf{p}}_i=(\hat{x}_i,\hat{y}_i)$ and ground-truth coordinate $\mathbf{p}_i=(x_i,y_i)$, we report coordinate hit rate and mean L2 error:
\begin{equation}
\mathrm{HIT}_{\mathrm{coord}}
=
\frac{1}{N_{\mathrm{p}}}
\sum_{i=1}^{N_{\mathrm{p}}}
\mathbf{1}\!\left[
\|\hat{\mathbf{p}}_i-\mathbf{p}_i\|_2 \leq \delta_i
\right],
\quad
\mathrm{L2}_{\mathrm{coord}}
=
\frac{1}{N_{\mathrm{p}}}
\sum_{i=1}^{N_{\mathrm{p}}}
\|\hat{\mathbf{p}}_i-\mathbf{p}_i\|_2 .
\end{equation}
where $\delta_i$ is the hit tolerance.

For trajectory prediction, with $\hat{\tau}_i=\{\hat{\mathbf{p}}_{i,t}\}_{t=1}^{T}$ and $\tau_i=\{\mathbf{p}_{i,t}\}_{t=1}^{T}$, we report trajectory hit rate:
\begin{equation}
\mathrm{HIT}_{\mathrm{traj}}
=
\frac{1}{N_{\tau}}
\sum_{i=1}^{N_{\tau}}
\mathbf{1}\!\left[
\frac{1}{T}\sum_{t=1}^{T}
\|\hat{\mathbf{p}}_{i,t}-\mathbf{p}_{i,t}\|_2
\leq \eta_i
\right],
\end{equation}
where $\eta_i$ is the hit tolerance. The default is $1.0$~m; \cref{tab:trajectory_tolerances} additionally reports $2.0$ and $5.0$~m tolerances for the same future-trajectory task. We also report trajectory mean L2 error.

\begin{table}[t]
\centering
\caption{Trajectory reasoning at multiple tolerances over 5 seconds. Hit rates are percentages; mean L2 is in meters. FT denotes fine-tuning. Unparseable outputs count as misses; -- indicates no valid numerical trajectory output.}
\label{tab:trajectory_tolerances}
\small
\setlength{\tabcolsep}{7pt}
\begin{tabular}{lccccc}
\toprule
Model & FT & Mean L2 $\downarrow$ & Hit@1m $\uparrow$ & Hit@2m $\uparrow$ & Hit@5m $\uparrow$ \\
\midrule
Qwen3-VL-8B & No & 138.76 & 0.0 & 0.0 & 0.2 \\
Qwen3-VL-8B & Yes & 19.90 & 0.0 & 1.7 & 9.7 \\
Qwen3.5-4B & No & -- & 0.0 & 0.0 & 0.0 \\
Qwen3.5-4B & Yes & 17.36 & 0.0 & 1.4 & 14.9 \\
Qwen3.5-9B & No & 425.68 & 0.0 & 0.0 & 0.0 \\
Qwen3.5-9B & Yes & \textbf{16.32} & \textbf{0.2} & \textbf{2.6} & \textbf{18.5} \\
\bottomrule
\end{tabular}
\end{table}

\paragraph{Trajectory calibration.}
For future object motion, each hit tests the mean displacement over all ten future waypoints spanning five seconds, not only the endpoint. Invalid trajectories count as misses; mean L2 is computed only when a valid numerical trajectory exists, so L2 and hit rates must be read together. \cref{tab:trajectory_tolerances} retains the original 1~m criterion and adds relaxed tolerances. Metric coordinates in the geometry benchmark may instead use normalized image units, so their L2 values are not directly comparable to trajectory L2 in meters.

\textbf{Text metrics.}
For free-form driving rationales in \texttt{driving\_reasoning\_trace}, token-level F1 and ROUGE-L are used as auxiliary reference-based measures:
\begin{equation}
\mathrm{R\text{-}L}
=
\frac{1}{N_{\mathrm{t}}}
\sum_{i=1}^{N_{\mathrm{t}}}
\mathrm{ROUGE-L}(\hat{s}_i, s_i),
\end{equation}
where $\hat{s}_i$ and $s_i$ are the predicted and reference textual answers. These text metrics are not treated as exact semantic equivalence measures; the main conclusions rely on structured choice, numerical, coordinate, and trajectory metrics.

\textbf{Tolerance configuration.}
The tolerances $\epsilon_i$, $\delta_i$, $\eta_i$ are stored per question in the dataset and follow a unified scheme. For numerical scalars,
\begin{equation}
\epsilon_i = \alpha_i + \rho \cdot |v_i|,\quad \rho = 0.05,
\end{equation}
where $\alpha_i$ is a subcategory-specific absolute floor and $\rho$ is a $5\%$ relative slack. \cref{tab:tolerance} lists the values used in our experiments.

\begin{table}[ht]
\centering
\small
\caption{Tolerance values used in numerical accuracy and hit-rate metrics.}
\label{tab:tolerance}
\begin{tabular}{l l c l}
\toprule
\textbf{Metric} & \textbf{Subcategory} & \textbf{Unit} & \textbf{Tolerance} \\
\midrule
$\mathrm{ACC}_{\mathrm{num}}$
  & \texttt{nearby\_lane\_count}            & count & $\epsilon = 0$ (exact) \\
  & \texttt{nearby\_crosswalk\_count}       & count & $\epsilon = 0$ (exact) \\
  & \texttt{object\_distance}               & m     & $\alpha = 1.0,\ \rho = 0.05$ \\
  & \texttt{object\_speed}                  & m/s   & $\alpha = 1.0,\ \rho = 0.05$ \\
  & \texttt{future\_time\_to\_conflict\_s}  & s     & $\alpha = 0.5,\ \rho = 0.05$ \\
\midrule
$\mathrm{HIT}_{\mathrm{coord}}$
  & \texttt{bbox\_center\_coordinates\_1000}    & px (0-1000) & $\delta = 8$ \\
  & \texttt{multiview\_center\_coordinates\_1000}& px (0-1000) & $\delta = 8$ \\
  & \texttt{object\_xy\_position}               & m            & $\delta = 1.0$ \\
\midrule
$\mathrm{HIT}_{\mathrm{traj}}$
  & \texttt{future\_xy\_trajectory}                  & m            & $\eta \in \{1.0,2.0,5.0\}$ \\
\bottomrule
\end{tabular}
\end{table}

\section{Additional Results}
\label{app:additional_vlm_reasoning}

This section reports additional evaluation results for nuReasoning. Unless otherwise stated, models are evaluated with multi-frame visual input when available. FT denotes fine-tuning on nuReasoning. Choice Acc. denotes choice accuracy, Num. Acc. denotes numerical accuracy within tolerance, Coord. Hit denotes coordinate hit rate, Traj. Hit denotes trajectory hit rate, and F1 denotes token-level F1 score. All accuracy, hit-rate, F1 score, and ROUGE-L values are reported as percentages. L2 errors are reported in task-specific coordinate units.

\subsection{Overall Answer-Format Performance}

\cref{tab:app_overall_answer_metrics} summarizes performance across answer formats. Fine-tuning produces large and consistent gains across categorical, numerical, coordinate, and free-form textual outputs. The greatest improvements occur for structured grounding. For example, Qwen3-VL-8B improves from $0.1\%$ to $46.3\%$ coordinate hit rate after fine-tuning, while Qwen3.5-9B improves from $0.3\%$ to $52.2\%$. These gains are accompanied by much lower coordinate L2 error.

The proprietary Gemini models achieve stronger zero-shot choice accuracy than most open-weight base models, but they remain weak on structured metric outputs. Both Gemini models obtain less than $3\%$ coordinate hit rate and $0.0\%$ trajectory hit rate. This suggests that strong general VLM ability does not automatically translate into precise spatial grounding for autonomous-driving reasoning.

Trajectory prediction remains the most challenging setting. Fine-tuning substantially reduces trajectory L2 error, but the strict trajectory hit rate remains close to zero. This indicates that trajectory reasoning is not solved by answer-format alignment and remains a useful stress test for future work.

\begin{table*}[ht]
\centering
\small
\renewcommand{\arraystretch}{1.08}
\setlength{\tabcolsep}{1.2pt}
\caption{Overall answer-format and structured-output metrics. Percentages are reported for accuracies, hit rates, token F1, and ROUGE-L. L2 values are reported in task-specific coordinate units. Coord.\ L2 pools ego-frame position errors (meters) and normalized image-coordinate errors (0-1000 scale), so it is comparable across rows but is not a physical
distance; Traj.\ L2 is in meters. FT denotes models fine-tuned on the nuReasoning training set.}
\label{tab:app_overall_answer_metrics}
\resizebox{\textwidth}{!}{
\begin{tabular}{l|cccccccc}
\toprule
Model 
& Choice Acc. $\uparrow$ 
& Num. Acc. $\uparrow$ 
& Coord. Hit $\uparrow$ 
& Coord. L2 $\downarrow$ 
& Traj. Hit $\uparrow$ 
& Traj. L2 $\downarrow$ 
& F1 $\uparrow$ 
& ROUGE-L $\uparrow$ \\
\midrule
Gemini-3.1-Pro        & 57.2 & 5.9  & 2.4  & 149.80 & 0.0 & 38.16  & 19.7 & 16.0 \\
Gemini-3-Flash        & 58.4 & 5.4  & 2.9  & 169.74 & 0.0 & 46.00  & 20.0 & 15.3 \\
\midrule
Qwen3-VL-8B           & 49.9 & 4.9  & 0.1  & 305.04 & 0.0 & 138.76 & 14.1 & 14.6 \\
Cosmos-Reason2-8B     & 49.3 & 5.3  & 0.0  & 298.45 & 0.0 & 77.39  & 18.9 & 15.9 \\
\rowcolor{blue!10}
Qwen3-VL-8B (FT)      & 85.4 & 34.4 & 46.3 & 31.38  & 0.0 & 19.90  & 41.9 & 40.7 \\
\midrule
Qwen3.5-4B            & 44.2 & 3.4  & 0.1  & 451.27 & 0.0 & --     & 20.1 & 16.5 \\
\rowcolor{blue!10}
Qwen3.5-4B (FT)       & \textbf{87.3} & 33.9 & 46.4 & 25.18  & 0.0 & 17.36  & \textbf{42.9} & \textbf{41.2} \\
\midrule
Qwen3.5-9B            & 52.4 & 4.7  & 0.3  & 212.17 & 0.0 & 425.68 & 19.5 & 18.3 \\
\rowcolor{blue!10}
Qwen3.5-9B (FT)       & 87.1 & \textbf{34.5} & \textbf{52.2} & \textbf{19.53} & \textbf{0.2} & \textbf{16.32} & 42.6 & 40.8 \\
\bottomrule
\end{tabular}
}
\end{table*}

\subsection{Representative Difficult Subcategories}

\cref{tab:app_hard_subcategories} compares several difficult subcategories across proprietary VLMs, open-weight base models, Cosmos-Reason2-8B, and fine-tuned Qwen3-VL-8B. The selected subcategories cover geometry, motion, driving decisions, and counterfactual risk assessment.

The results show a clear distinction between answer selection and metric-grounded reasoning. Gemini-3.1-Pro and Gemini-3-Flash are competitive on some choice-style tasks, but both perform poorly on coordinate grounding. For instance, on \texttt{bbox\_center\_coordinates\_1000}, Gemini-3.1-Pro obtains $3.7\%$ coordinate hit rate and Gemini-3-Flash obtains $4.7\%$, while Qwen3-VL-8B (FT) reaches $57.5\%$. A similar pattern appears for ego-frame object position, where Qwen3-VL-8B (FT) reaches $31.3\%$ while all base models remain below $1\%$.

Numerical geometry also benefits strongly from fine-tuning. On \texttt{object\_distance}, Qwen3-VL-8B improves from $10.0\%$ to $74.9\%$. Counterfactual risk assessment shows another large gain, increasing from $44.7\%$ to $86.5\%$. By contrast, future trajectory prediction remains unsolved under the strict hit-rate metric, with all models at $0.0\%$ on the selected trajectory task.

\begin{table*}[ht]
\centering
\renewcommand{\arraystretch}{1.08}
\setlength{\tabcolsep}{1.5pt}
\caption{Representative difficult subcategories. Values are percentages. Each row reports the metric used for that subcategory.}
\label{tab:app_hard_subcategories}
\resizebox{\textwidth}{!}{
\begin{tabular}{l|l|l|c|cccc}
\toprule
Category 
& Subcategory 
& Metric 
& Count 
& Gemini-3.1-Pro 
& Cosmos-Reason2-8B 
& Qwen3-VL-8B 
& Qwen3-VL-8B (FT) \\
\midrule
Geometry 
& \texttt{object\_xy\_position} 
& Coord. Hit $\uparrow$ 
& 466 
& 0.6 & 0.0 & 0.0 & \textbf{31.3} \\

Geometry 
& \texttt{bbox\_center\_coordinates\_1000} 
& Coord. Hit $\uparrow$ 
& 617 
& 3.7 & 0.0 & 0.2 & \textbf{57.5} \\

Geometry 
& \texttt{object\_distance} 
& Num. Acc. $\uparrow$ 
& 451 
& 14.4 & 8.9 & 10.0 & \textbf{74.9} \\

Motion 
& \texttt{future\_xy\_trajectory} 
& Traj. Hit $\uparrow$ 
& 422 
& 0.0 & 0.0 & 0.0 & 0.0 \\

Motion 
& \texttt{object\_speed} 
& Num. Acc. $\uparrow$ 
& 330 
& 16.1 & 16.4 & 14.5 & \textbf{31.5} \\

Motion 
& \texttt{future\_time\_to\_conflict\_s} 
& Num. Acc. $\uparrow$ 
& 198 
& 6.1 & 7.6 & 7.1 & \textbf{16.2} \\

Driving 
& \texttt{driving\_decision\_joint} 
& Choice Acc. $\uparrow$ 
& 1,017 
& 44.4 & 38.8 & 41.0 & \textbf{58.4} \\

Counterfactual 
& \texttt{action\_risk\_assessment} 
& Choice Acc. $\uparrow$ 
& 3,051 
& 37.7 & 46.5 & 44.7 & \textbf{86.5} \\
\bottomrule
\end{tabular}
}
\end{table*}

\begin{table*}[ht]
\centering
\renewcommand{\arraystretch}{1.1}
\setlength{\tabcolsep}{5pt}
\caption{Subcategory-level comparison for Qwen3-VL-8B before and after fine-tuning under the default multi-frame setting. Values are percentages. Each row uses the metric appropriate for that subcategory. The best result in each row is bolded.}
\label{tab:app_qwenvl_subcategory_breakdown}
\resizebox{\textwidth}{!}{
\begin{tabular}{l|l|l|c|cc}
\toprule
Capability 
& Subcategory 
& Metric 
& Count 
& Before FT 
& After FT \\
\midrule
\multirow{15}{*}{Geometry}
& \texttt{object\_xy\_position} 
& Coord. Hit $\uparrow$ 
& 466 
& 0.0 
& \textbf{31.3} \\

& \texttt{object\_relative\_position} 
& Choice Acc. $\uparrow$ 
& 699 
& 33.5 
& \textbf{99.6} \\

& \texttt{object\_distance} 
& Num. Acc. $\uparrow$ 
& 451 
& 10.0 
& \textbf{74.9} \\

& \texttt{distance\_from\_2d\_bbox} 
& Choice Acc. $\uparrow$ 
& 713 
& 30.7 
& \textbf{71.8} \\

& \texttt{bbox\_center\_coordinates\_1000} 
& Coord. Hit $\uparrow$ 
& 617 
& 0.2 
& \textbf{57.5} \\

& \texttt{normalized\_bbox\_coordinates\_1000} 
& Num. Acc. $\uparrow$ 
& 588 
& {0.0} 
& {0.0} \\

& \texttt{ego\_position\_from\_2d\_bbox} 
& Choice Acc. $\uparrow$ 
& 645 
& 40.2 
& \textbf{99.2} \\

& \texttt{image\_region\_from\_3d\_object} 
& Choice Acc. $\uparrow$ 
& 723 
& 38.5 
& \textbf{90.7} \\

& \texttt{camera\_view\_for\_3d\_object} 
& Choice Acc. $\uparrow$ 
& 580 
& 45.2 
& \textbf{95.7} \\

& \texttt{closest\_object\_type} 
& Choice Acc. $\uparrow$ 
& 162 
& 80.2 
& \textbf{98.8} \\

& \texttt{multiview\_camera\_pair} 
& Choice Acc. $\uparrow$ 
& 287 
& 64.8 
& \textbf{99.0} \\

& \texttt{current\_lane\_coordinates\_2d\_1000} 
& Num. Acc. $\uparrow$ 
& 173 
& {0.0} 
& {0.0} \\

& \texttt{nearby\_lane\_count} 
& Num. Acc. $\uparrow$ 
& 167 
& 0.0 
& \textbf{98.2} \\

& \texttt{nearby\_crosswalk\_count} 
& Num. Acc. $\uparrow$ 
& 160 
& 1.2 
& \textbf{76.9} \\

& \texttt{nearby\_crosswalk\_coordinates\_2d\_1000} 
& Num. Acc. $\uparrow$ 
& 147 
& {0.0} 
& {0.0} \\
\midrule
\multirow{7}{*}{Motion}
& \texttt{future\_motion\_label} 
& Choice Acc. $\uparrow$ 
& 401 
& 38.2 
& \textbf{57.4} \\

& \texttt{future\_xy\_trajectory} 
& Traj. Hit $\uparrow$ 
& 422 
& {0.0} 
& {0.0} \\

& \texttt{future\_path\_intersection} 
& Choice Acc. $\uparrow$ 
& 503 
& 97.6 
& \textbf{99.0} \\

& \texttt{future\_time\_to\_conflict\_s} 
& Num. Acc. $\uparrow$ 
& 198 
& 7.1 
& \textbf{16.2} \\

& \texttt{conflict\_prediction} 
& Choice Acc. $\uparrow$ 
& 764 
& 99.6 
& \textbf{100.0} \\

& \texttt{motion\_relation} 
& Choice Acc. $\uparrow$ 
& 554 
& 34.5 
& \textbf{100.0} \\

& \texttt{object\_speed} 
& Num. Acc. $\uparrow$ 
& 330 
& 14.5 
& \textbf{31.5} \\
\midrule
\multirow{4}{*}{Driving}
& \texttt{driving\_decision\_joint} 
& Choice Acc. $\uparrow$ 
& 1,017 
& 41.0 
& \textbf{58.4} \\

& \texttt{driving\_reasoning\_trace} 
& ROUGE-L $\uparrow$ 
& 1,017 
& 14.6 
& \textbf{40.7} \\

& \texttt{traffic\_regulation\_light\_state} 
& Choice Acc. $\uparrow$ 
& 488 
& 81.6 
& \textbf{88.7} \\

& \texttt{traffic\_regulation\_signal} 
& Choice Acc. $\uparrow$ 
& 97 
& 45.4 
& \textbf{97.9} \\
\midrule
\multirow{2}{*}{Counterfactual}
& \texttt{unsafe\_action\_identification} 
& Choice Acc. $\uparrow$ 
& 1,017 
& 44.5 
& \textbf{66.9} \\

& \texttt{action\_risk\_assessment} 
& Choice Acc. $\uparrow$ 
& 3,051 
& 44.7 
& \textbf{86.5} \\
\bottomrule
\end{tabular}
}
\end{table*}

\subsection{Subcategory-Level Effect of Fine-Tuning}

\cref{tab:app_qwenvl_subcategory_breakdown} provides a more detailed before-and-after comparison for Qwen3-VL-8B. The gains are broad rather than concentrated in a single task family. Fine-tuning improves relative-position reasoning, distance estimation, position inference from 2D boxes, camera-view reasoning, nearby-lane counting, traffic signal recognition, motion-relation reasoning, and counterfactual risk assessment.

The largest gains are observed on tasks with constrained structured outputs. For example, \texttt{object\_relative\_position} increases from $33.5\%$ to $99.6\%$, \texttt{ego\_position\_from\_2d\_bbox} increases from $40.2\%$ to $99.2\%$, and \texttt{motion\_relation} increases from $34.5\%$ to $100.0\%$. These improvements indicate that the nuReasoning dataset helps the model map visual evidence into the required reasoning schema.

However, several dense coordinate-output tasks remain difficult. In particular, \texttt{normalized\_bbox\_coordinates\_1000}, \texttt{current\_lane\_coordinates\_2d\_1000}, and \texttt{nearby\_crosswalk\_coordinates\_2d\_1000} remain at $0.0\%$ after fine-tuning. These failures show that fine-tuning improves many structured reasoning skills, but precise normalized coordinate generation and lane/crosswalk geometry still require further modeling improvements.

\subsection{Three-Seed Planning Ablation}
\label{sec:supp_planning_analysis}
\label{app:planning_seeds}
\label{app:reasoning_controls}
\cref{tab:planning_three_seeds} reports mean and standard deviation across three random seeds, separately from the single-run results in \cref{vla_results,tab:controls_transfer}. All variants share the controlled protocol in Appendix~\cref{app:baseline_protocol}. Full S+D+CF supervision improves mean NPS by 8.34 points over planning only. Spatial supervision produces the largest progress gain, from 87.30 to 91.19. Decision supervision increases NC from 94.44 to 96.48 and DA from 91.67 to 94.74. CF increases mean NPS from 71.51 to 72.33 when added to D, and from 71.19 to 72.86 when added to S+D.

Adding D to S improves NC, DA, and NPS but lowers HL from $61.70\pm0.52$ to $60.08\pm0.65$. High-level action/rationale supervision does not uniquely determine the maneuver's timing, speed profile, or lateral placement; a multi-task trade-off with exact trajectory imitation is a plausible explanation rather than a proven mechanism. The full model attains the highest mean DA, HL, and NPS. Three-seed variation reflects repeated training, not a confidence interval over unseen environments or a statistical test of every pairwise comparison.

\begin{table}[htbp]
\centering
\caption{nuVLA planning ablation across three random seeds (mean $\pm$ standard deviation). All variants receive planning supervision; Comf denotes comfort.}
\label{tab:planning_three_seeds}
\footnotesize
\setlength{\tabcolsep}{4pt}
\begin{tabular}{lcccccc}
\toprule
Reasoning & NC $\uparrow$ & DA $\uparrow$ & EP $\uparrow$ & Comf. $\uparrow$ & HL $\uparrow$ & NPS $\uparrow$ \\
\midrule
None & $94.44\pm0.15$ & $91.67\pm0.30$ & $87.30\pm0.23$ & $99.77\pm0.11$ & $55.14\pm0.30$ & $64.52\pm0.31$ \\
D & $96.48\pm0.30$ & $94.74\pm0.35$ & $89.89\pm0.34$ & $99.93\pm0.06$ & $61.48\pm0.66$ & $71.51\pm0.41$ \\
D+CF & $96.12\pm0.30$ & $95.89\pm0.21$ & $90.32\pm0.25$ & $99.87\pm0.06$ & $61.41\pm0.23$ & $72.33\pm0.06$ \\
S & $94.93\pm0.35$ & $93.52\pm0.28$ & $91.19\pm0.30$ & $99.74\pm0.11$ & $61.70\pm0.52$ & $70.18\pm0.21$ \\
S+D & $96.22\pm0.79$ & $95.26\pm0.43$ & $90.22\pm0.28$ & $99.77\pm0.11$ & $60.08\pm0.65$ & $71.19\pm0.34$ \\
S+D+CF & $96.25\pm0.17$ & $96.08\pm0.06$ & $90.55\pm0.27$ & $99.87\pm0.06$ & $61.95\pm0.28$ & $\mathbf{72.86\pm0.07}$ \\
\bottomrule
\end{tabular}
\end{table}

\subsection{Reasoning-Content and Inference Controls}
\label{app:inference_control}

\paragraph{Conditioning on generated reasoning.}
We train an additional decision-reasoning variant that generates only the meta-action and reasoning trace. Its action expert attends jointly to prompt-side VLM features and hidden states of the generated tokens. The comparison model is standard nuVLA trained with decision reasoning only, which plans directly from prompt-side features. Thus, this is a comparison of trained conditioning designs, not a generation switch applied to the full S+D+CF checkpoint. Runtime is measured on an NVIDIA RTX Pro 6000 GPU.

\begin{table}[htbp]
\centering
\caption{Decision-supervised nuVLA with and without generated test-time reasoning. Latency is measured on the same GPU; lower is better.}
\label{tab:test_time_reasoning}
\small
\setlength{\tabcolsep}{12pt}
\begin{tabular}{lccc}
\toprule
Generated reasoning & NPS $\uparrow$ & HL $\uparrow$ & Latency (ms) $\downarrow$ \\
\midrule
No (prompt-side features) & 70.91 & 61.01 & 202.10 \\
Yes (prompt + generated tokens) & 70.23 & 60.41 & 1711.91 \\
\bottomrule
\end{tabular}
\end{table}

The generated-reasoning variant has similar, slightly lower NPS/HL and approximately $8.5\times$ latency. This supports retaining the training-time benefit without autoregressive reasoning at test time in this setting, not a universal conclusion about test-time reasoning. We do not interpret lexical trace quality as a reliable scalar of semantic correctness or report sample-level trace-trajectory correlations, which would also be confounded by scene difficulty.

\subsection{Sensitivity to Work-Zone Training Proportion}
\label{app:workzone_control}
We randomly downsample work-zone clips until they represent 20\% of training data, retaining all non-work-zone clips. Removed clips and their reasoning/VQA samples are not replaced; validation and test data remain unchanged. We fine-tune Qwen3.5-4B on the corresponding QA subset and train nuVLA (S+D+CF) on the corresponding clips under otherwise identical settings. This is a \emph{reduced-data} experiment: it changes both category proportions and supervision volume, and cannot isolate balance at fixed training-set size.

\begin{table}[t]
\centering
\caption{Qwen3.5-4B reasoning after capping work-zone training clips at 20\%. All values are percentages. Evaluation subsets are unchanged.}
\label{tab:workzone_reasoning}
\small
\setlength{\tabcolsep}{4pt}
\begin{tabular}{llccccc}
\toprule
Training fraction & Evaluation & Geometry Ch. & Coord. Hit & Motion Ch. & Driving Ch. & CF Ch. \\
\midrule
Original (42.8\%) & Overall & 92.5 & 46.4 & 92.2 & 78.4 & 83.3 \\
Capped (20.0\%) & Overall & 89.7 & 34.3 & 91.9 & 78.3 & 82.1 \\
Original (42.8\%) & Work-zone & 93.1 & 46.6 & 92.6 & 79.4 & 80.3 \\
Capped (20.0\%) & Work-zone & 90.2 & 35.2 & 92.3 & 79.1 & 79.1 \\
Original (42.8\%) & Non-work-zone & 91.8 & 46.0 & 91.6 & 77.0 & 87.5 \\
Capped (20.0\%) & Non-work-zone & 88.9 & 33.0 & 91.2 & 77.1 & 86.2 \\
\bottomrule
\end{tabular}
\end{table}

\begin{table}[htbp]
\centering
\caption{nuVLA (S+D+CF) overall planning after limiting work-zone training clips to 20\%. Removed clips are not replaced, so training-set size is not held fixed.}
\label{tab:workzone_planning}
\small
\setlength{\tabcolsep}{18pt}
\begin{tabular}{lc}
\toprule
Training work-zone fraction & Overall NPS $\uparrow$ \\
\midrule
Original (42.8\%) & 73.09 \\
Capped (20.0\%) & 68.12 \\
\bottomrule
\end{tabular}
\end{table}

Downsampling does not consistently improve reasoning on non-work-zone samples: coordinate hit rate decreases from 46.0\% to 33.0\%, while driving choice accuracy changes from 77.0\% to 77.1\%. Overall planning NPS decreases from 73.09 to 68.12. These results provide no evidence of a benefit from this particular cap, but do not establish the optimal work-zone ratio or rule out benefits from fixed-budget rebalancing or additional non-work-zone data.

\subsection{Qualitative Results}

\begin{figure}[ht]
    \centering
    \includegraphics[width=\linewidth]{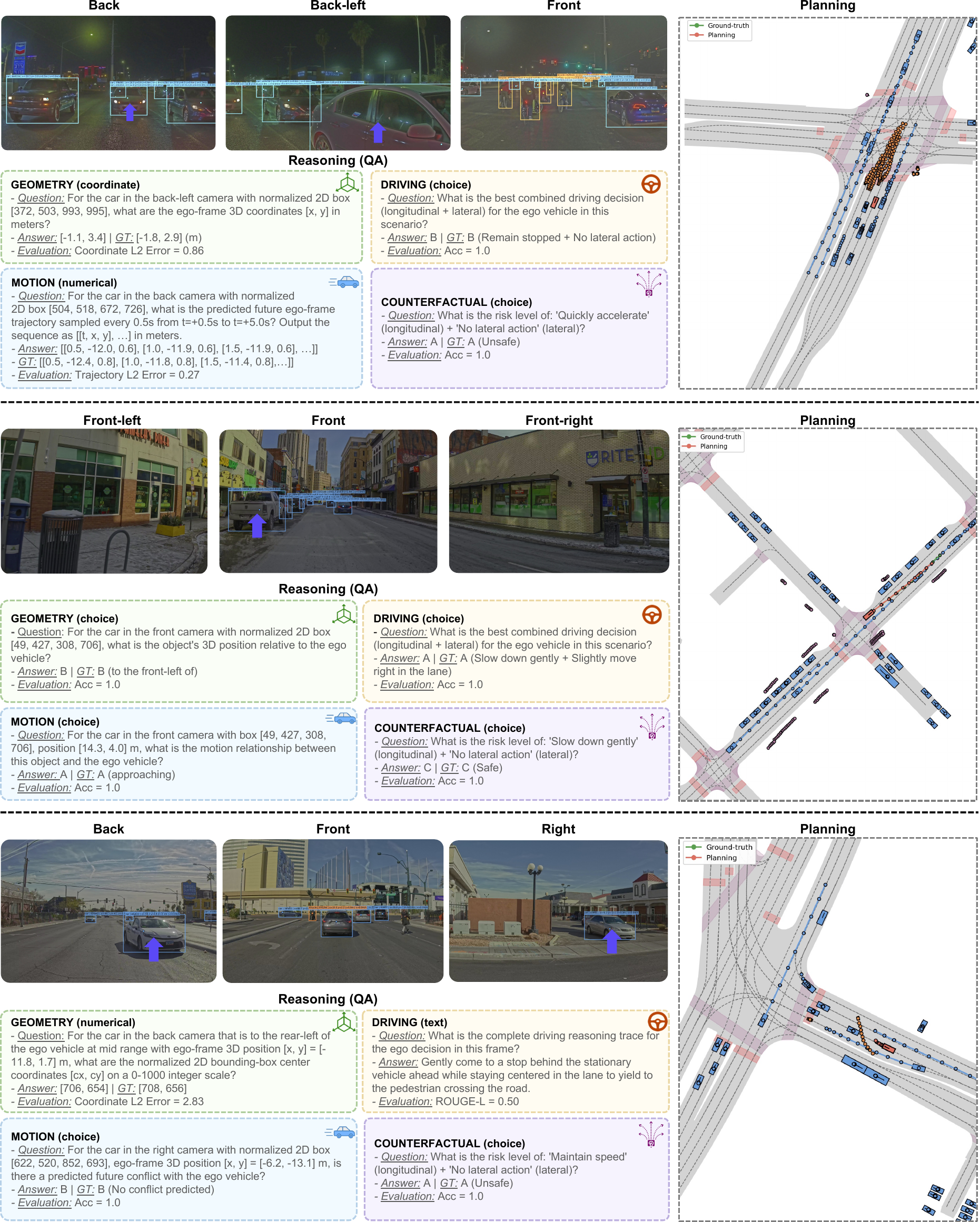}
    \caption{Additional qualitative results of reasoning and planning performance. Reasoning is evaluated using the fine-tuned Qwen3.5-9B model, while planning is evaluated using the nuVLA model trained with all types of reasoning supervision (S+D+CF).}
    \label{fig:planning}
    \vspace{-0.3cm}
\end{figure}

We provide additional qualitative reasoning and planning results in \cref{fig:planning}. These examples illustrate individual outputs rather than category-level success rates. As the lane-change explanation in \cref{results} shows, a plausible action and substantial lexical overlap can coexist with disagreement about the dominant constraint; text similarity should therefore be interpreted alongside structured targets and trajectory metrics.

\section{Broader Impact and Safeguards}
\label{app:broader_impact_safeguards}

\textbf{Broader impact.}
nuReasoning aims to support research on safer autonomous driving in challenging long-tail scenarios. Its spatial, decision, and counterfactual reasoning annotations can help models better understand scene context, anticipate risks, and distinguish safe, unsafe, and suboptimal actions. The benchmark also enables systematic evaluation of reasoning and its effect on planning.

However, nuReasoning is an offline benchmark and is not sufficient evidence for real-world deployment. Models may still fail under unseen cities, weather, sensor setups, traffic rules, or rare scenarios outside the dataset, and may overfit to benchmark-specific open-loop metrics. Systems developed with nuReasoning should therefore undergo additional simulation, safety analysis, and real-world validation before deployment.

\textbf{Safeguards.}
Because nuReasoning is built from real-world AV fleet logs, it may contain privacy-sensitive visual information. Before release, we apply privacy-preserving anonymization, including blurring faces and license plates. We also use automated filtering and human expert verification to remove unsuitable samples and improve annotation quality. The dataset is intended for research and benchmark use under the accompanying license, usage terms, and privacy and safety requirements.